\documentclass{article}
\usepackage{natbib}
\usepackage{jfrExamplee}
\usepackage{graphicx}
\usepackage{apalike}
\usepackage{setspace}
\usepackage{epstopdf}
\usepackage{todonotes}
\usepackage{algorithm}
\usepackage{verbatim} \usepackage{caption}
\usepackage{subcaption}
\usepackage{tabulary}
\usepackage{tabularx}
\usepackage{makecell}

\usepackage{mathtools}
\usepackage{textcomp}
\usepackage{multirow}
\usepackage{amssymb}  \usepackage{url}
\usepackage{units}
\usepackage{enumitem}
\usepackage{adjustbox}
\usepackage{rotating}
\usepackage{booktabs}
\usepackage{xcolor,colortbl}
\usepackage[section]{placeins}
\usepackage[hidelinks]{hyperref}

\usepackage{pgfplots}

\newcommand{\vecx}{\mathbf{x}}
\newcommand{\mean}{\boldsymbol{\mu}}
\newcommand{\variance}{\boldsymbol{\Sigma}}

\definecolor{light_green}{rgb}{0.77,0.93,0.8}
\definecolor{light_yellow}{rgb}{1.0,0.93,0.61}
\definecolor{light_red}{rgb}{1.0,0.78,0.8}

\newcommand *\naive {Na\"{i}ve }

\title{
A Sweet Pepper Harvesting Robot for Protected Cropping Environments
}
\author{
Chris Lehnert
\thanks{ Direct correspondence to: Chris Lehnert: c.lehnert@qut.edu.au } \\
Electrical Engineering and Computer Science\\
Queensland University of Technology\\
Brisbane 4000, Australia \\
\texttt{c.lehnert@qut.edu.au} \\
\And
Chris McCool \\
Electrical Engineering and Computer Science\\
Queensland University of Technology\\
Brisbane 4000, Australia \\
\texttt{c.mccool@qut.edu.au} \\
\And
Inkyu Sa \\
Mechanical and Process Engineering\\
ETH Zurich\\
Zurich 8092, Switzerland \\
\texttt{inkyu.sa@mavt.ethz.ch}
\And
Tristan Perez \\
Electrical Engineering and Computer Science\\
Queensland University of Technology\\
Brisbane 4000, Australia \\
\texttt{t.perez@qut.edu.au} \\
}


\begin{document}

\setlength{\textfloatsep}{20pt}

\maketitle

\begin{abstract}
Using robots to harvest sweet peppers in protected cropping environments has remained unsolved despite considerable effort by the research community over several decades.
In this paper, we present the robotic harvester, Harvey, designed for sweet peppers in protected cropping environments that achieved a 76.5\% success rate (within a modified scenario) which improves upon our prior work which achieved 58\% and related sweet pepper harvesting work which achieved 33\%. This improvement was primarily achieved through the introduction of a novel peduncle segmentation system using an efficient deep convolutional neural network, in conjunction with 3D post-filtering to detect the critical cutting location. We benchmark the peduncle segmentation against prior art demonstrating a considerable improvement in performance with an $F_1$ score of 0.564 compared to 0.302. The robotic harvester uses a perception pipeline to detect a target sweet pepper and an appropriate grasp and cutting pose used to determine the trajectory of a multi-modal harvesting tool to grasp the sweet pepper and cut it from the plant. A novel decoupling mechanism enables the gripping and cutting operations to be performed independently. We perform an in-depth analysis of the full robotic harvesting system to highlight bottlenecks and failure points that future work could address. 

\end{abstract}
\section{Introduction}\label{sec:introduction}
The horticulture industry is heavily reliant on manual labour.
For instance, in Australia, labour hire is between 20\% and 30\% of total cash costs~\citep{ABARE2014}. 
These costs along with other pressures such as high cost of inputs (energy, water, agrochemicals, \textit{etc}.), variable production due to uncertain weather events and labour scarcity are putting profit margins for horticulture farms under tremendous pressure.

Robotic harvesting offers a potentially attractive solution by not only reducing costs of labour but by lowering risks associated with obtaining labour and food safety. Robot harvesting also enables capitalising on opportunities for extended selective harvesting that maximises quality - optimal scheduling for harvesting different parts of the farm with required quality thresholds. For these reasons, there has been increasing interest in the use of agricultural robots for harvesting crop and vegetables over the past three decades \citep{Kondoetall2011}. The task of developing a robotic harvester is particularly challenging and requires the integration of numerous subsystems such as crop detection, motion planning, and dexterous manipulation. The underlying functional requirements share those of manufacturing, but there are additional challenges: uncontrolled and changing lighting, variability in crop size and shape, occlusions, and the delicate nature of the crop being manipulated. A survey of robotic harvesting of horticulture crops reviewed 50 projects over the past 30 years \citep{Bac2014}. The review highlights that over this period the performance of automated harvesting has not improved substantially despite advances in sensors, computers, and machine intelligence. If robotic-crop harvesting is to become a reality, two key challenges must be addressed:

\begin{enumerate}[topsep=0pt,itemsep=-1ex,partopsep=1ex,parsep=1ex]
	\item perception of the crop and environment, and
	\item manipulation of the crop. 
\end{enumerate}

Perception relates to being able to locate or segment the crop, determine its location in 3D and locate key points for attaching and detaching the crop from the plant.
Crop manipulation involves attaching and detaching the crop without harming the crop or plant; this involves the development of physically appropriate end-effectors combined with algorithms to effectively and efficiently utilise them.
Both perception and crop manipulation are challenging tasks due to the presence of occluding obstacles such as leaves and branches, as well as natural variability in crop size, shape, and pose.

\begin{figure}[tbh]
	\centering
	\includegraphics[width=0.8\columnwidth]{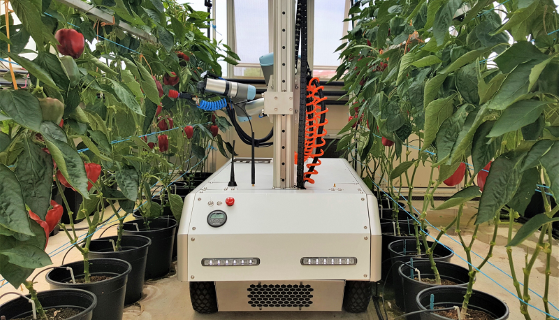}
		\caption{The Harvey platform, an autonomous sweet pepper harvester operating in a protected cropping system.}
	\label{fig:frontpage1}
\end{figure}

In this paper, we present a new robotic harvester (Harvey) designed for sweet peppers (also known as capsicum or bell pepper) in protected cropping environments that improves upon prior work~\citep{Lehnert2017}.
In principle, the robotic harvester uses a perception pipeline to detect a target sweet pepper and an appropriate grasp and cutting pose used to determine the trajectory of a multi-modal harvesting tool. The harvesting tool features a suction cup to grasp the sweet pepper and an oscillating blade to cut the pepper from the plant. A novel decoupling mechanism enables the gripping and cutting operations to be performed serially with independently chosen grasping and cutting trajectories. This combination of robotic-vision techniques and crop manipulation tools are key enabling factors for the successful harvesting of high-value crops, in particular, sweet peppers. Fig~\ref{fig:frontpage1} shows an example of the crop setup and characteristics as expected in a protected cropping environment.

This work improves upon~\citep{Lehnert2017} through the introduction of an accurate peduncle segmentation system (in the perception pipeline) and improved integration of the perception to action system. We perform an in-depth analysis of the full robotic harvesting system to highlight bottlenecks and failure points that future work could address. These improvements in the perception and action methods considerably increase the harvesting success rate from 58\% to 76.5\%, under a modified scenario in a real protected cropping system. Analysis of the full robotic harvesting system highlights that the integration of an active vision method would likely improve both sweet pepper and peduncle segmentation within highly occluded scenarios.

The presented harvesting system achieved a 76.5\% success rate (within a modified scenario) which improves upon our prior work which achieved 58\% and related sweet pepper harvesting work which achieved 33\%~\citep{Bac2017} (which has some differences in the cropping system)

Central to the improved harvesting performance of Harvey, is the novel use of an efficient deep convolutional neural network~\citep{McCool17_1}, referred to as \textit{MiniInception}, in conjunction with 3D post-filtering to detect the cutting location (for sweet peppers this is the peduncle---the part of the fruit which is attached to the plant). We benchmark the \textit{MiniInception} approach for peduncle segmentation against prior art~\citep{Sa17_1} demonstrating a considerable improvement in performance with an $F_1$ score of 0.564 compared to 0.302. This improvement is possible not only due to the increased accuracy of the deep convolutional neural network but also the novel use of 3D post-filtering.

The key contributions of this paper are:
\begin{itemize}
	\item A proven in-field robotic harvesting system that achieves a harvesting success rate of 76.5\% in a modified scenario,
	\item an in-depth analysis of the perception and harvesting field trials of the robotic harvester,
	\item and a novel method for peduncle segmentation using an efficient deep convolutional neural network in conjunction with 3D post-filtering.
\end{itemize}

The remainder of the paper is structured as follows. A review of the current state of the art methods for autonomous harvesting of horticultural crops is presented in Section \ref{sec:literature}. The design of the autonomous harvesting platform is then presented in Section \ref{sec:system_design}, outlining the harvesting environment, platform and design of the multi-modal end-effector tool. The methods for perception and planning are then presented in Section \ref{sec:perception_and_planning}, outlining our novel techniques for segmentation and 3D localisation of sweet peppers. Results of three experiments are presented in Section \ref{sec:exp_results}, presenting the performance of our segmentation and peduncle localisation methods. The last experiment presents the results for our end-to-end autonomous harvesting system in a real protected cropping environment. Section \ref{sec:discussion_and_conclusion} discusses key challenges and future work for improving the performance of autonomous harvesting systems for horticulture.

\section{Literature}\label{sec:literature}

Current literature contains examples of various robots which are capable of autonomous harvesting under certain environmental conditions and crops including: sweet peppers \citep{Bac2017} including our previous work \citep{Lehnert2017}, cucumbers \citep{Henten2002}, citrus \citep{Mehta2014}, strawberries \citep{Hayashi2010} and apples \citep{Bulanon2010a,De-An2011}.
Despite this, the commercial application of such robots for horticulture is very limited.
Some of the factors behind this lack of commercial uptake, as reviewed in \citet{Bac2014} and \citet{Shamshiri2018}, include the complexity of agricultural environments and the different configuration of crops within it (poses, sizes, shapes and colours). 
In addition, the highly occluded nature of the scene combined with the requirements of high efficiency, accuracy, and robustness of the manipulation process has led to very few systems being commercially viable.
The above factors are the subject of great attention in the literature recently and can be divided into two categories: perception and manipulation. 

The most related to our work is the sweet pepper harvesting robot developed within the Clever Robots for Crops (CROPS) project~\citep{Bac2017,Hemming2014a,Bontsema2014}. This robot was developed for harvesting sweet peppers using a 9DOF manipulator within a greenhouse environment. In this work, a colour and time of flight camera are used in an eye-in-hand configuration to detect and localise the crop. Using depth information the position of the sweet peppers and orientation of the stem are estimated. In the work of \citet{Bac2017}, two different end effector designs where field tested where the best design was shown to achieve a harvesting success of 6\% in an unmodified crop. This result led the developers to simplify the crop configuration by removing crop clusters and occluding leaves. This led to an improvement in harvesting success of the robot up to 33\% for the simplified scenario. An average harvesting time per sweet pepper was reported as 94 seconds in the work of \citet{Bac2017}. 

\subsection{Perception}

Crop perception includes detection, segmentation and 3D localisation, and has been investigated for a variety of different crops. The key challenges include detection and segmentation in challenging outdoor environments. 
3D Crop localisation refers to the process of determining the position and orientation information of the crop~\citep{VanHenten2003,Kitamura2008,Hemming2014a,Bulanon2010}
One of the challenges with localisation includes fusing multiple modalities of sensing technology such as visual (colour or texture) information with depth information to obtain an accurate 3D localisation of the crop. 

\subsection{Detection and Segmentation}

For detection and segmentation, using standard RGB cameras, researchers have explored the use of either traditional features or deep learnt features. 
Examples of traditional features include the use of a radial symmetry transform to perform grape detection~\citep{Nuske_2011_6891,Nuske:2014aa}, the detection of a distinctive specular reflective pattern to detect apples in \citet{Wang_2012_7240} or combining colour and shape features to perform semantic segmentation, into four classes, for tomato detection in \citet{Yamamoto:2014aa}.

More recently, feature learning approaches have been explored.
One of the earliest examples of this was in 2013 where \citet{Hung:2013aa} proposed to learn features using an auto-encoder.
These features were then used within a conditional random field (CRF) framework to perform almond segmentation.
This approach achieved impressive segmentation performance but did not perform object detection. 

In 2016, \citet{Sa16_1} proposed the use of deep learning systems for detection of nine different crops (e.g. sweet pepper, melons, apples and avocados) and explored different methods for combining multi-modal information (i.e., early, or late fusion of multispectral images) and explored some of the limits of such an approach.
For crop counting, \citet{Rahnemoonfar16_1} proposed to learn deep convolutional neural networks using simulated training data to count apples.

The above methods have addressed issues such as crop segmentation~\citep{Hung:2013aa}, detection~\citep{Sa16_1} or estimating the number of crops in a sub-region of the image~\citep{Rahnemoonfar16_1}.
However, to perform harvesting, it is important to find other attributes of a plant such as the peduncle; this is the part of the fruit which attaches it to the stem or branch of the plant.

In terms of peduncle detection, \citet{cubero2014} demonstrated the detection of various fruit peduncles using radius and curvature signatures. 
The Euclidean distance and the angle rate of change between each of the points on the contour and the fruit centroid
are calculated. The presence of peduncles yields rapid changes in these metrics and can be detected using a specified threshold. 
\citet{blasco2003machine} and \citet{Ruiz:1996aa} presented peduncle detection of oranges, peaches, and apples using a Bayesian discriminant model of RGB colour information. 
The size of a colour segmented area was calculated and assigned to pre-defined classes. 
The above methods are more likely suitable for the quality control and inspection of crop peduncles after the crop has been harvested rather than for harvesting automation as they require an inspection chamber that provides ideal lighting conditions with a clean background, no occlusions, good viewpoints, and high-quality static imagery.

In our previous work, \citet{Sa17_1} proposed the use of point feature histograms and colour features to detect sweet pepper peduncles.
This approach was evaluated on data from a real-world cropping environment and achieved impressive results.
A downside of this approach was the requirement to annotate 3D imagery, which can be time-consuming.
An alternative approach would be to make use of just the 2D imagery and employ a deep convolutional neural network (DCNNs).

Recent work has demonstrated the potential for the use of deep learning approaches to address agricultural computer vision problems.
\citet{McCool17_1} proposed an approach for deploying efficient deep convolutional neural networks for crop vs weed classification by distilling the information from a high-performance but high computational load neural network to efficient smaller, \textit{student}, networks. Semantic segmentation making use of synthetic imagery was proposed by \citet{BARTH2017} for plant-part segmentation and \citet{milioto2018arxiv} presented an efficient framework for semantic segmentation of weeds.

Given the success of the above work~\citep{McCool17_1}, we considered its application in this paper and describe it in more detail in Section~\ref{sec:peduncle_detection}. We note that at the time of the experiments the prior work~\citep{BARTH2017} and~\citep{milioto2018arxiv} were unavailable.

\subsubsection{Crop Localisation}

In most cases of the literature, a vision system is used to detect and segment the target crop, and depth information is used to determine its position. Methods for estimating depth include the use of stereo images \citep{VanHenten2003,Kitamura2008}, time-of-flight cameras \citep{Hemming2014a} and laser range finders \citep{Bulanon2010}. 

Colour and depth sensors have been used by \citet{Nguyen2014} to segment bushels of apples using Euclidean clustering techniques. Furthermore, random sample consensus was used to fit a spherical model to each apple in order to estimate their centroids.

In some cases, the orientation of the crop is estimated for use in grasping and detachment stages of the harvesting process~\citep{Henten2002,Han2012}. For instance, suction cup grippers (commonly used in harvesting) have the disadvantage of failing if there is no complete seal on the crop. Estimating the orientation of an asymmetrical crop such as sweet peppers or strawberries can aid in the alignment of the suction cup gripper improving the attachment success rates \citep{Hayashi2010,Lehnert2016}. 

Studies on 3D crop localisation in the presence of occlusions have been shown to improve localisation accuracy, such as the work by \citet{Gongal2015}. Partially visible crops are highly challenging to localise since only a portion of information is available. To address this issue, using spacial or temporal multiple-views and their registration technologies has been utilised. \citet{gongal2016apple} employed a dual-sided imaging system that consists of 5 pairs of colour and 3D cameras for each side (10 pairs in total) for apple localisation.

\subsection{Motion Planning}

A survey of autonomous harvesting projects for horticulture by \citet{Bac2014} has found that more than half of the reviewed projects do not report the motion planning techniques used and this can account for the lack of progress in motion planning techniques for horticulture. Two standard methods for motion planning include open loop planning and visual servoing. 

Open loop planning methods which do not simultaneously localise the crop and plan the motion is a common approach for autonomous harvesting. Open loop methods can suffer from problems when the robot interacts with the scene inducing changes to the crop location and thereby reducing the accuracy of the current estimate. If the robot only interacts minimally with the scene and if the scene is static, open loop planning methods can be successful \citep{Hemming2014a,Baur2014,VanHenten2003,Scarfe2000}. Other improvements over standard motion planners have been attempted, such as using optimal path planning to determine the best motion of the manipulator for harvesting crop seen in the work by \citet{Schuetz2015}.

Image-based visual servoing has been used to control the motion of a robot manipulator to a target crop using an eye-in-hand camera for autonomous harvesting of citrus~\citep{Mehta2014,Hannan2004}, apples~\citep{Baeten2008,Bulanon2010}, tomatoes~\citep{Kondo1996} and strawberries~\citep{Han2012,Hayashi2010}, but also with a fixed point of view camera for sweet peppers~\citep{Kitamura2005}. The approach by \citet{Mehta2014} uses a perspective transformation to estimate the position of the crop in Euclidean space to determine the control policy of the manipulator. Visual servoing can be useful for motion planning within dense vegetation where crop localisation can perform poorly due to occlusions~\citep{Barth2016}.

Successfully grasping and detachment in a dense and cluttered environment is still an ongoing research problem and is currently an active area of research \citep{Bhattacharjee2014,Jain2013,Killpack2015}, often requiring tactile sensing to discern between rigid and deformable objects. As advocated in \citet{Bac2014}, simplifying the workspace or developing harvesting tools which simplify the harvesting operation are potential solutions to the motion planning problem in dense and cluttered horticultural environments.

\subsubsection{Harvesting Tools and Manipulators}

The most common manipulator used for autonomous harvesting over the past 50 years has been 3DOF cartesian and anthropomorphic arms, followed by 6DOF manipulators \citep{Bac2014}. Optimal design of manipulators for different horticulture tasks such as cucumbers \citep{VanHenten2009} and sweet peppers \citep{Lehnert2015} has been used to aid in the selection of the joint type and number of DOF for the manipulator. This work has shown that a potential optimal design for harvesting within a sweet pepper environment is a 6DOF manipulator with two extra linear DOF at the base adding vertical and horizontal freedom \citep{Lehnert2015}. 

A variety of different harvesting tools have been developed for grasping and manipulating crop. These range from suction cups, contact grippers, soft robotic fingers and under-actuated anthropomorphic grippers. One of the most widely used gripper technologies to handle crop is based on suction cups~\citep{Blanes2011} and the use of vacuum pressure to grasp the crop. Suction cups have been used for a large range of crops such as tomatoes~\citep{Ling2005}, apples~\citep{Baeten2008}, cucumbers~\citep{VanHenten2003}, sweet peppers~\citep{Hemming2014a,Bontsema2014} and strawberries~\citep{Hayashi2010}. A suction cup has the advantage of requiring less workspace to complete the grasp (i.e. the mechanism is not required to envelop the crop but only come in contact with a smaller patch of its surface.)

Contact grippers use friction to hold onto a crop, where the most common is a two finger jaw gripper~\citep{Monkman2007}. Contact-based grippers which use mostly two or three fingers have been used for harvesting apples~\citep{Bulanon2010,De-An2011}, tomatoes~\citep{Ling2005}, oranges~\citep{De-An2011}, and kiwifruit~\citep{Scarfe2000}. Using soft robotic fingers instead of rigid fingers has shown to have the potential for crop harvesting~\citep{SoftRobotics2016,Ilievski2011} as they have the advantage of reducing grasping damage via compliant and soft interaction with the crop. However, the soft fingers can be difficult to implement and require further refinement before they can be used practically.

Often within horticulture, a detachment tool is required to remove the crop from the plant. These often require specific designs for the target crop, depending on how the crop is attached to the plant. The most common type of detachment tool is a mechanism that cuts or severs the peduncle of the crop and include thermal cutters, scissors or a custom cutting mechanism.

Thermal cutters have been used for a variety of crops such as sweet peppers~\citep{Bachche2013} and cucumbers~\citep{Henten2002} which have the advantage of sealing the cut area preventing the spread of diseases. Scissor type cutters have been used for different crops such as sweet peppers~\citep{Kitamura2005,Hemming2014a} and strawberries~\citep{Hayashi2010,Han2012}. In these approaches, the gripper and the cutter are mounted at fixed offsets with each other and require the crop and peduncle to fit within the fixed offset for a successful detachment. An added DOF between the gripper and cutter can be included to tackle this problem~\citep{Kondo2010}, but can add to the complexity of the end effector design. A disadvantage of a scissor mechanism is the potential to damage surrounding parts of the plant~\citep{Hemming2013b}. Furthermore, scissors cut in a plane, and if the curvature of the peduncle or stem is irregular, then the cutting plane may not completely sever it.

A custom cutting tool for sweet peppers was developed in~\citep{Hemming2014a} which used the concept of enveloping the sweet pepper with a hinged jaw mechanism. This type of mechanism was more successful than a scissor mechanism developed within the same project. The main disadvantage with this mechanism was the size and geometry constraints required to get the mechanism around the back of the sweet pepper. This was problematic as the mechanism would get stuck on parts of the plant surrounding the sweet pepper~\citep{Hemming2013b}.

\section{System Design}\label{sec:system_design}

\begin{figure}[tbh]
	\includegraphics[width=\columnwidth]{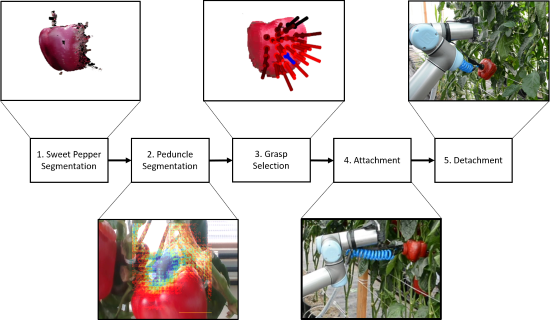}
	\centering
	\caption{The five stages of the autonomous harvesting cycle. The robot firstly detects a target pepper at a wide viewing angle and then moves to a close-range perspective. The sweet pepper is then segmented using colour information. Thirdly the peduncle of the sweet pepper is estimated using a deep learning segmentation method. The fourth stage determines the optimal grasp location for attachment. The final stage uses the estimated peduncle and grasp pose to execute the attachment and detachment tools to remove the sweet pepper from the plant.}
	\label{fig:system_stages}
\end{figure}

This section describes the system design for our autonomous sweet pepper harvester and includes an overview of the robotic platform, harvesting tool, software design and a description of the harvesting environment. The overall procedure for harvesting sweet peppers is shown in Fig~\ref{fig:system_stages} and can be broken down into five stages:
\begin{enumerate}[topsep=0pt,itemsep=-1ex,partopsep=1ex,parsep=1ex]
	\item Sweet Pepper Segmentation
	\item Peduncle Segmentation
	\item Grasp Selection
	\item Attachment
	\item Detachment 
\end{enumerate}

The first three steps: Sweet pepper localisation, peduncle localisation and grasp selection form the perception system described in Section \ref{sec:perception_and_planning}. During the sweet pepper localisation stage, the robot arm is moved to a long-range perspective to capture a 3D colour image of the whole scene using an eye-in-hand RGB-D camera. A target sweet pepper is localised at the long-range perspective and used to move the camera to a close-range perspective of the targeted sweet pepper in order to improve the performance of the peduncle localisation. The next stage localises the peduncle of the target sweet pepper using a convolutional neural network and a 3D filtering method to estimate the centroid of the peduncle. A grasp selection is then performed on the segmented 3D points of the targeted sweet pepper. The grasp selection uses a heuristic to rank possible grasp poses on the target sweet pepper using surface and position metrics.  

The harvesting method involves two stages, crop attachment and crop detachment. The harvesting method uses a custom harvesting tool described in Section \ref{sec:harvest_tool} comprised of a suction cup to grasp the sweet pepper (attachment), and an oscillating blade to cut the sweet pepper from the plant (detachment). The grasp pose and peduncle pose from the previous perception system are used to plan the motion of the robot arm. The grasp pose is used to plan the attachment stage whereas the peduncle pose is used to plan the detachment. During the attachment step, the suction cup is magnetically attached to the cutting blade. For the subsequent detachment phase, the cutting blade is separated from the suction cup which remains attached to the end-effector via a flexible tether. This design enables the robot arm to perform the attachment and detachment steps sequentially at independently chosen locations (grasp and peduncle poses) to maximise the success of both phases.

\subsection{Protected Cropping Environment} \label{sec:cropping_environment}

\begin{figure}[tbh]
	\centering
	\begin{subfigure}[b]{0.49\textwidth}
		\centering
		\label{fig:greenhouse_image}
		\includegraphics[height=6cm]{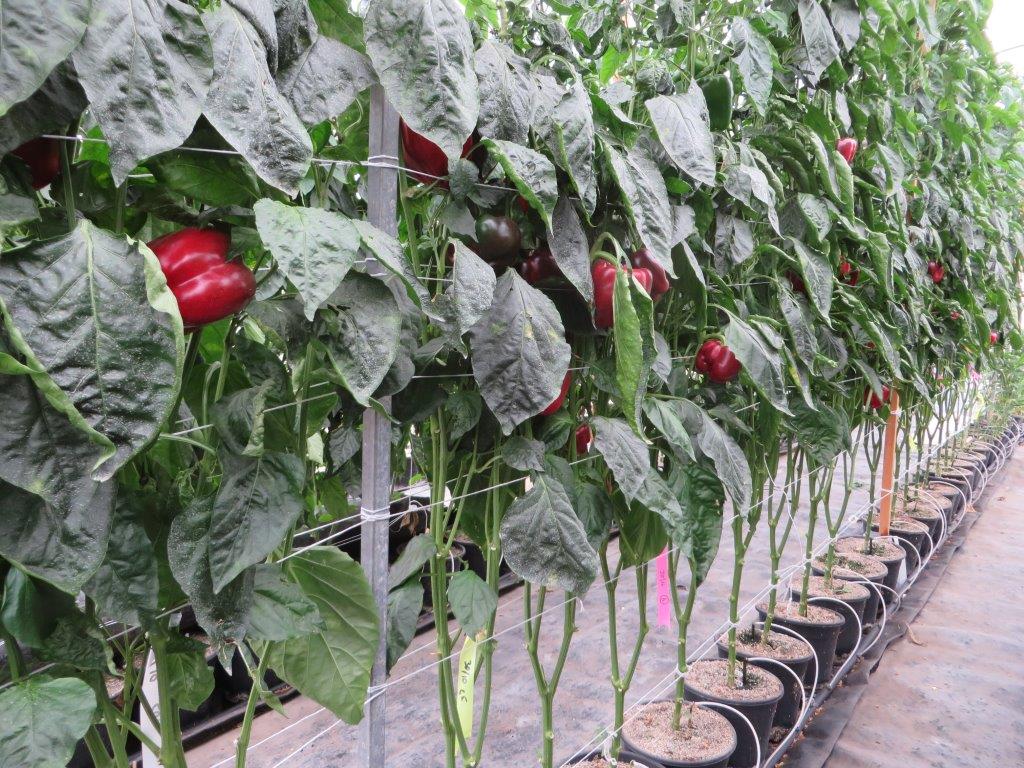}
		\caption{}
	\end{subfigure}
	\begin{subfigure}[b]{0.49\textwidth}
		\centering
		\label{fig:greenhouse_diagram}
		\includegraphics[height=6cm]{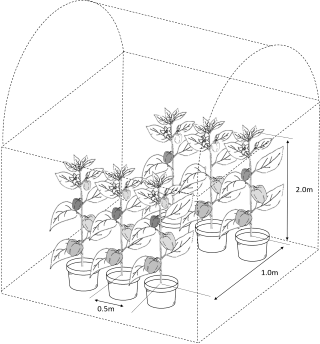}
		\caption{}
	\end{subfigure}
	\caption{(a) A section of a crop row in a protective cropping environment. Plants are trained onto a trellis so that the crop is presented on a two dimensional planar surface. Some occlusions from leaves can be seen. (b) Plant layout within protected cropping facility including typical row and plant spacing.}
	\label{fig:greenhouse_diagram_image}
\end{figure}

Commercially, sweet peppers are grown in both outdoor field and indoor protected cropping environments. 
In this work, we focus on the task of picking sweet peppers within a protected cropping environment.
Protected cropping environments comprise of glass or a semi-permanent plastic enclosure (poly-tunnel) designed to prevent damage to the crop from pests, heat, cold, rain and wind. Protected cropping systems in tropical climates such as Northern Australia can differ to other international greenhouse systems with respect to trellising and potting methods. However, the underlying plant structure such as leaves, stems and sweet peppers are very similar, including their physical and visual appearance.

Within the enclosure, sweet peppers are typically grown hydroponically with support trellises or wiring, allowing them to grow up to \unit[4]{m} tall in a relatively 2D planar structure. 

Protected cropping environments are designed to provide significantly increased yields compared to field-grown crops. The layout also provides three significant advantages for autonomous harvesting. First, the crop is presented on a two-dimensional planar surface. This planar structure significantly reduces occlusion from other branches and leaves, making visually detecting and locating crops much easier than field sweet peppers that grow within a low, three-dimensional bush/shrub. Second, the planar presentation of the crop simplifies collision avoidance and motion planning by providing relatively open access to both the stem and side face of the crop. Third, the protective cropping environment presents a more forgiving environment for computer vision as the plastics and/or mesh roof and walls diffuse incoming sunlight. Diffused sunlight means the environment is lit relatively evenly, with virtually no sharp shadows.  

The protected cropping system that our studies were conducted on grew two cultivars of sweet pepper, Mercuno and Ducati with a row spacing of approximately \unit[1]{m} and a plant spacing of \unit[0.5]{m}. The layout of this system is illustrated in Fig~\ref{fig:greenhouse_diagram_image} along with a view of a typical section of the sweet pepper crop in a commercial protected cropping system.

\subsection{Platform Design}

\begin{figure}[tbh]
	\centering
	\includegraphics[height=7cm]{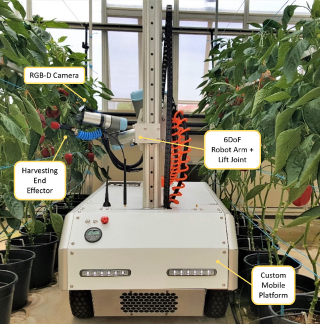}
	\includegraphics[height=7cm]{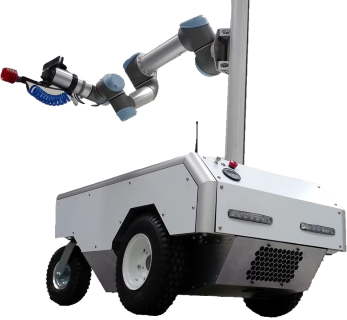}
	\caption{The Harvey platform with each component highlighted. The components consist of a 6DOF UR5 robot arm with a harvesting tool attached to its end effector, a custom mobile base platform with PC and control box and a prismatic lift joint.}
	\label{fig:harvey_platform}
\end{figure}

The harvesting robot ``Harvey'', is shown in Fig~\ref{fig:harvey_platform}. 
Harvey is comprised of a 7DOF manipulator, a custom end-effector and a mobile base that houses batteries, a water-cooled computer with dedicated graphics and the appropriate control hardware for operating the manipulator, end-effector and drive system. Fig~\ref{fig:harvey_platform} highlights each component, showing the setup of the robotic platform within the protected cropping facility. 

\begin{figure}[tbh]
	\centering
	\includegraphics[height=8cm]{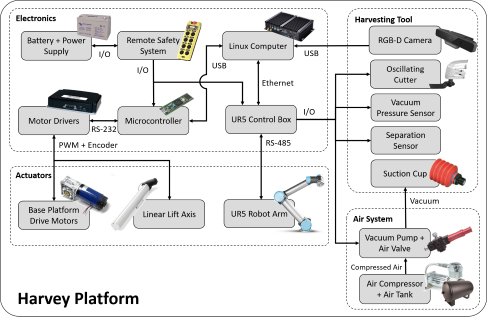}
	\caption{A high level schematic diagram of the autonomous harvesting ``Harvey" platform. The system is comprised of the electronics, actuators, harvesting tool and air system. The diagram illustrates how each component is interconnected to create the full autonomous platform.}
	\label{fig:schematic_diagram}
\end{figure}

Harvey was designed to work within a protected cropping environment and manoeuvre within each crop row which can grow up to \unit[2]{m} tall. The vertical lift axis and mobile base were selected specifically to allow the end-effector to reach all of the sweet pepper along each row. 

To choose a robot arm suitable for the sweet pepper harvesting task, a range of arms where compared with respect to their cost, base weight, payload, workspace, IP rating and speed. A range of industrial arms were surveyed including the Kuka LBR, Kinova Jaco/MicoA, Barrett WAM, Universal Robots (UR3/5/10) and Shunk Powerball. A UR5, from Universal Robots, was selected as it satisfied all of the specifications for the harvesting task. In particular, the UR5 is IP54 rated for water and dust, has a workspace area of 0.85m and a payload of \unit[5]{kg}, suitable for the weight of the harvesting tool (\unit[2]{kg}) and the weight of sweet pepper which can weigh up to \unit[0.5]{kg}.

\subsection{Harvesting Tool Design} \label{sec:harvest_tool}

The end-effector of the robotic arm is a custom designed harvesting tool. This is the means by which the robot interacts with the crop and its design is critical to reliable attachment and detachment of the sweet pepper.

\begin{figure}[tbh]
	\centering
	\includegraphics[height=6.25cm]{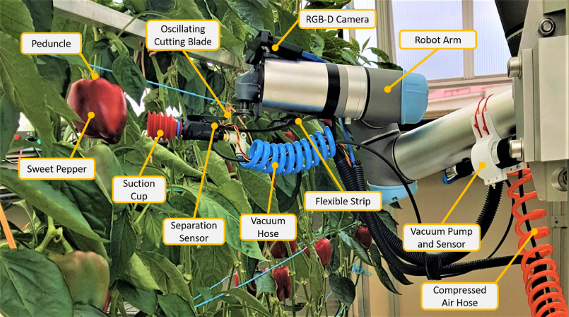}
	\centering
	\caption{Harvesting Tool. The harvesting tool attached to the end effector of the robot manipulator. The tool is comprised of a suction cup for grasping which can separate from a cutting tool (oscillating cutting blade) via a passive magnetic coupling. The suction cup is attached to a flexible rubber strip allowing the suction cup to freely move when separated from the cutting tool. The harvesting tool has a separation sensor as feedback to the system if the tool separates accidentally. A vacuum pressure sensor is also used to measure whether the suction cup is attached to a sweet pepper. A RGB-D camera is also used as input for the perception system (segmentation of sweet peppers and peduncles).}
	\label{fig:gripper_labelled}
\end{figure}

This harvesting tool performs a dual purpose, performing both gripping and cutting operations sequentially. This tool is designed to remove a sweet pepper by first gripping it with a vacuum gripper, then cutting through the stem with an oscillating blade. The individual components of the harvesting tool are mounted on the tool point of the robot arm and are shown in Fig~\ref{fig:gripper_labelled}. An RGB-D sensor is mounted near the front of the end effector body, and is used to identify the shape and location of each sweet pepper. The body of the end effector contains a hand-held oscillating multi-tool with metal blade for cutting stems. 

Grasping and cutting a sweet pepper at the same time with a single harvesting tool can be challenging. The natural variation of sweet peppers and peduncles (size, shape and orientation) imposes challenging constraints on the path of the end effector if attempting to grasp and cut at the same time. To overcome this difficulty, our end-effector has a decoupling mechanism (using passive magnets) which separates the grasping and cutting tools by a flexible tether. Separating the grasping and cutting tool relaxes the constraint for simultaneous grasping and cutting, enabling these operations to occur sequentially and at different locations (i.e not at fixed offsets from gripper and cutter). This allows a wider set of grasping and cutting poses to be used within the planning algorithm.
  
The decoupling mechanism is composed of two distinct components. The first component is a flat, flexible polymer strip that attaches the suction cup to the body of the end-effector. The second component is a magnet that allows the base of the suction cup to attach to the underside of the cutting blade (see Fig~\ref{fig:gripper_labelled}). 

During the gripping operation, the suction cup is magnetically attached to the cutting blade, allowing the robot arm to control the position of the suction cup to grasp the sweet pepper. During the cutting operation, the suction cup passively detaches from the cutting blade, while remaining attached to the body of the end effector by the flexible strip, allowing the suction cup to move independently of the cutting blade. 
This simple and passive decoupling method requires no additional active elements such as electronics, actuators or sensors, and allows independent gripping and cutting locations to be chosen for each sweet pepper, which in turn enables more reliable harvesting. 

Two feedback sensors are also integrated into the harvesting tool. The first is a separation sensor (SPDT lever switch) which indicates whether the suction cup and cutting tool are coupled or separated. The second is a vacuum pressure sensor which provides feedback on whether the suction cup has grasped a sweet pepper as the vacuum pressure is directly related to the holding force.

\begin{figure}[tbh!]
	\centering
	
	\begin{subfigure}[b]{0.32\columnwidth}
		\centering
		\label{fig:gripper_views:1}
		\includegraphics[height=3.25cm, trim={8cm 8cm 14cm 0cm},clip]{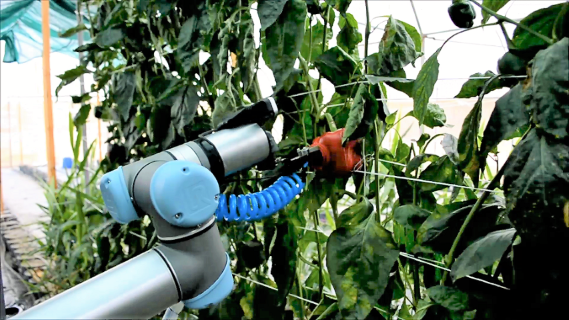}
		\caption{}
	\end{subfigure}
	\begin{subfigure}[b]{0.32\columnwidth}\centering
		\centering
		\label{fig:gripper_views:2}
		\includegraphics[height=3.25cm, trim={8cm 8cm 14cm 0cm},clip]{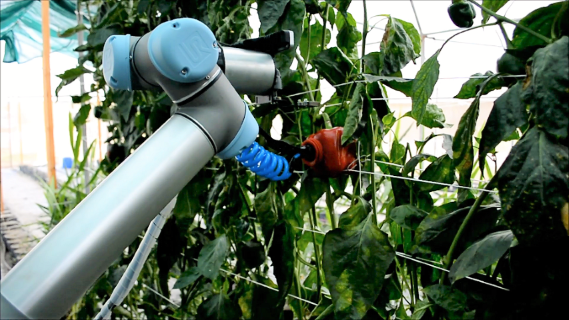}
		\caption{}
	\end{subfigure}
	\begin{subfigure}[b]{0.32\columnwidth}\centering
		\label{fig:gripper_views:3}
		\includegraphics[height=3.25cm, trim={8cm 8cm 14cm 0cm},clip]{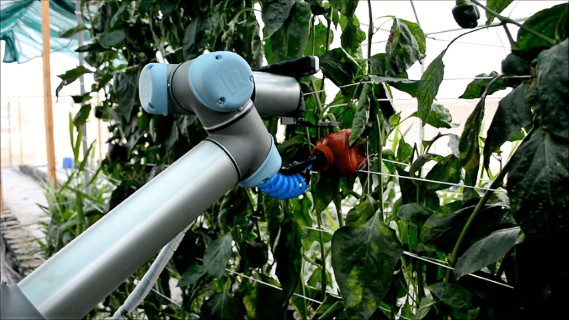}
		\caption{}
	\end{subfigure}
	
	\begin{subfigure}[b]{0.4\columnwidth}\centering
		\label{fig:gripper_views:4}
		\includegraphics[height=3.25cm, trim={8cm 6cm 14cm 2cm},clip]{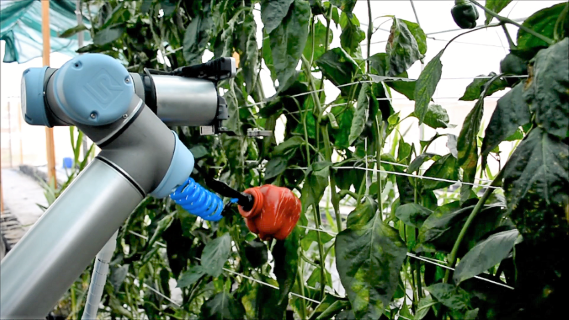}
		\caption{}
	\end{subfigure}
	\begin{subfigure}[b]{0.4\columnwidth}\centering
		\label{fig:gripper_views:5}
		\includegraphics[height=3.25cm]{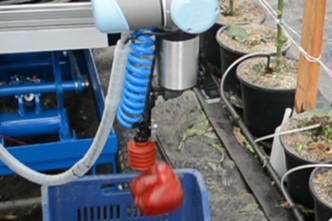}
		\caption{}
	\end{subfigure}
	\caption{Indicative images of the harvesting phases from attachment, separation of the cutter and detachment of the sweet pepper. (a) Image of the harvesting tool with the suction cup coupled to the tool during the attachment stage. (b) A vertical motion is used to decouple the suction cup and cutting tool. (c) Once separated, the cutting tool is used to sever the peduncle from the plant and is shown decoupled from the suction cup. (d) Post detachment, the sweet pepper remains attached to the suction cup but falls away from the plant. (e) As the sweet pepper is placed into a crate, the suction cup and cutting tool passively re-attach magnetically.}
	\label{fig:harvesting_overview1}
\end{figure}

The procedure for harvesting the sweet pepper with the harvesting tools, illustrated in Fig~\ref{fig:harvesting_overview1}, is as follows:
\begin{enumerate} [topsep=0pt,itemsep=-1ex,partopsep=1ex,parsep=1ex]
	\item Attach to crop: The arm is moved to allow the suction cup to attach to the surface of the sweet pepper.  The attachment point is chosen as a smooth flat area on the face of the sweet pepper (Fig~\ref{fig:harvesting_overview1}a). 
	\item separate suction cup from cutting blade: The end effector is moved upwards which causes the magnets holding the suction cup to the underside of the cutting blade to separate. The flexible strip now allows the cutting blade to move independently of the suction cup/sweet pepper so that the cutting blade can target the optimum stem-cutting location (Fig~\ref{fig:harvesting_overview1}b). 
	\item Peduncle cutting:  The oscillating cutting blade is moved forward to cut the sweet pepper peduncle, detaching it from the plant. After the peduncle is cut, the sweet pepper falls away from the plant whilst remaining attached to the suction cup, in-turn attached to the end-effector by the flexible strip (Fig~\ref{fig:harvesting_overview1}c).  
	\item Magnet re-attachment and Release crop: The robot arm is moved so the end effector points downwards over a collection crate. This passively re-attaches magnetically the suction cup with the cutting blade under the force of gravity, ready to harvest another sweet pepper. The vacuum is then released from the suction cup causing the crop to drop from a small height safely into the collection crate (Fig~\ref{fig:harvesting_overview1}e). 
\end{enumerate}

 \subsection{Software Design}\label{sec:software_design}

\begin{figure}[bh!]
	\centering
	\begin{subfigure}[b]{0.425\textwidth}\centering
				\raisebox{15mm}{\includegraphics[width=\columnwidth]{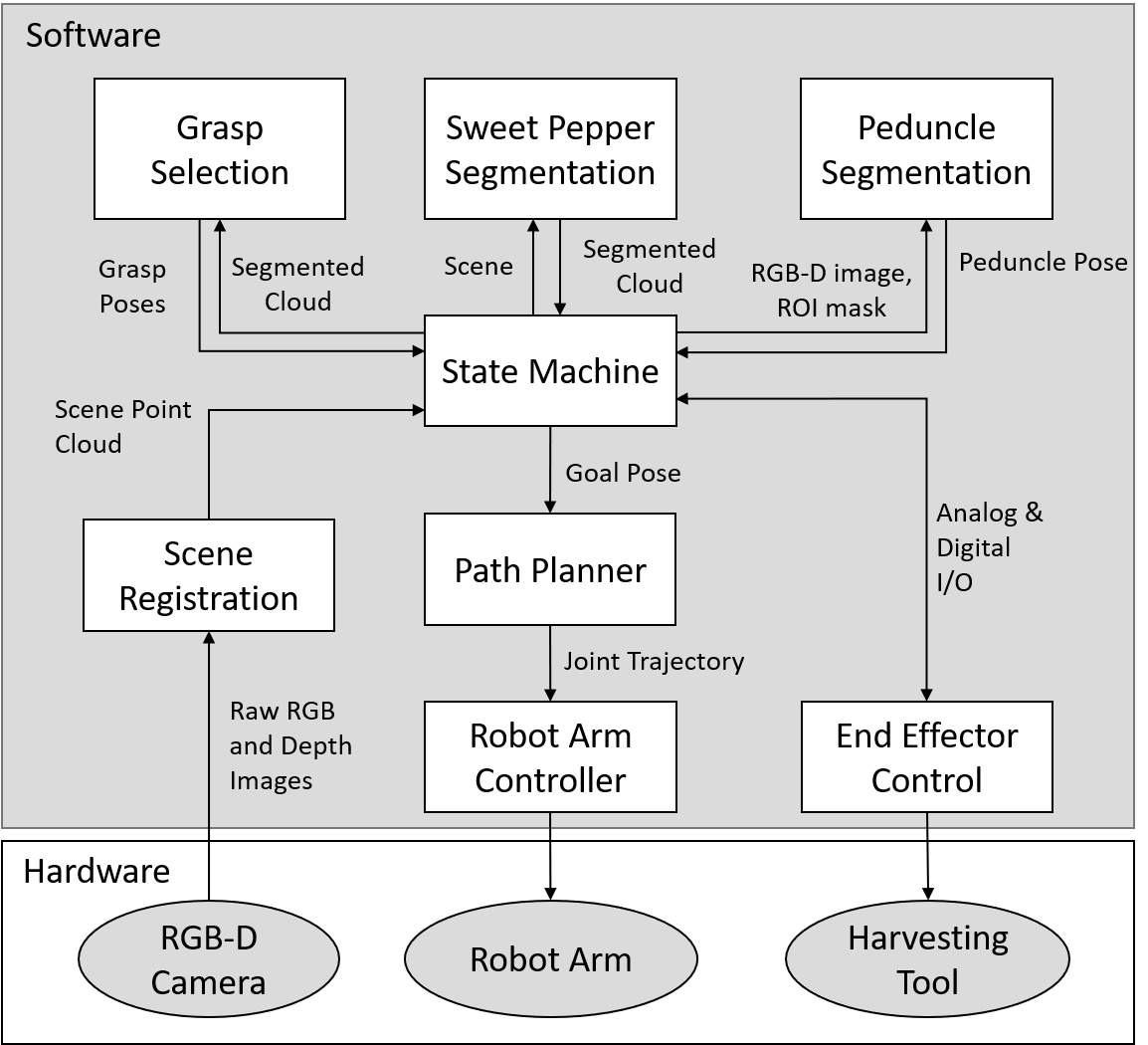}}
		\caption{}
	\end{subfigure}\quad
	\begin{subfigure}[b]{0.45\textwidth}\centering
				\raisebox{0mm}{\includegraphics[height=9cm]{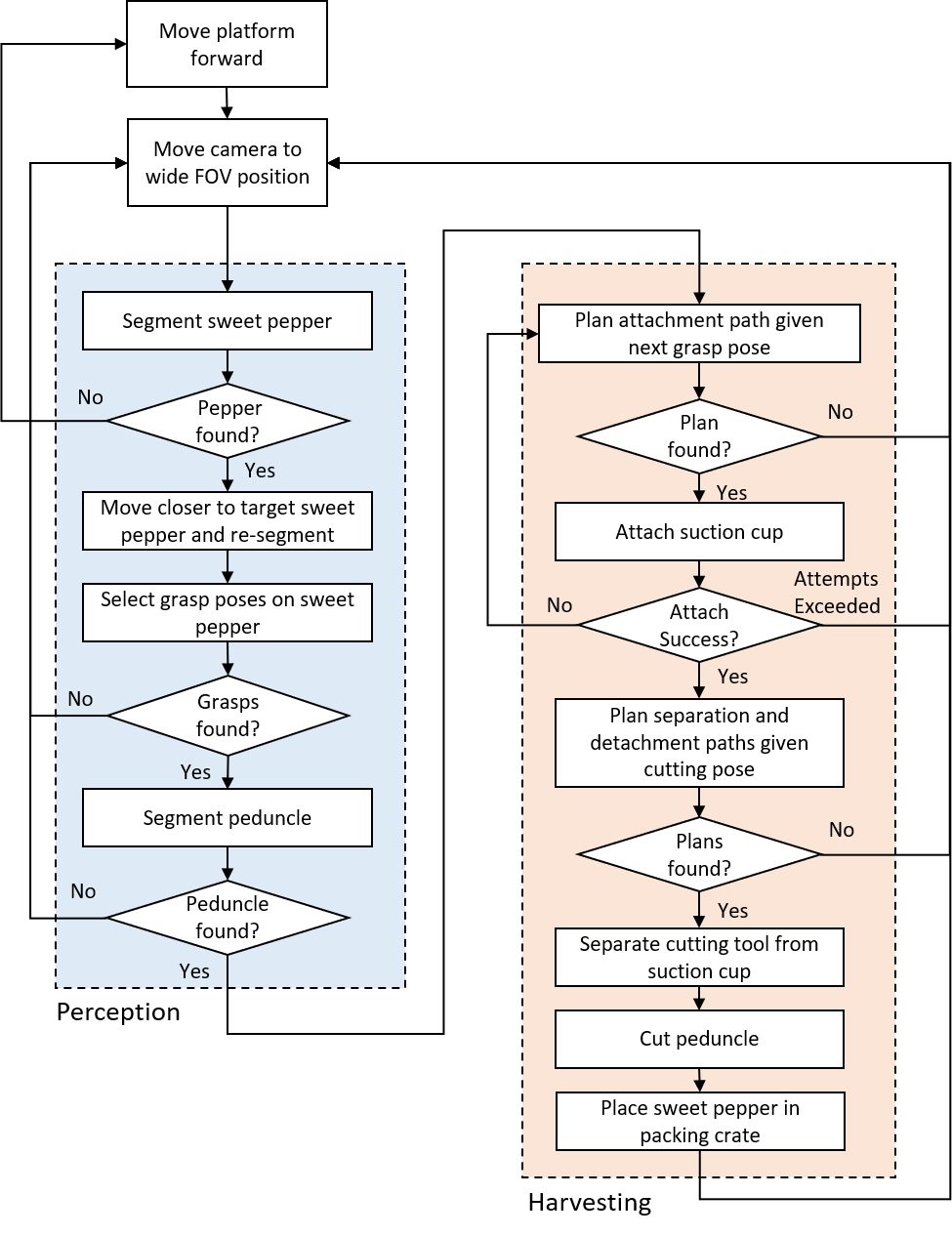}}
		\caption{}
	\end{subfigure}
	\caption{Software architecture diagrams (a) Diagram illustrating how each software subsystem is connected. The state machine is central to the Harvey system, and contains the decision making logic. The Scene Registration system reconstructs the 3D scene from RGB and depth images. The Sweet pepper segmentation, peduncle segmentation and grasp Selection  Fitter are used to generate information required to harvest each sweet pepper within a scanned scene. Physical harvesting operations are handled by the Path Planner, Robot Arm Controller and End Effector Control subsystems. (b) Logic flow diagram which is implemented within the state machine subsystem and illustrates the decision making steps for the autonomous harvesting operation}
	\label{fig:system_diagram}
\end{figure}

The software system was designed within the Robot Operating System (ROS) framework. The system contains eight customised ROS nodes as illustrated in Fig~\ref{fig:system_diagram}a. Communication among the nodes was primarily performed with custom messages sent by ROS actions and services. Central to the software design was a state machine implemented with the ROS SMACH package \citep{smach}.

Fig~\ref{fig:system_diagram}b shows the logic flow diagram which is internal to the state machine for harvesting sweet peppers along a crop row. At each stage the system checks if a failure occurred, such as if a sweet pepper is not detected, or a suction cup attachment failed. If a failure occurred with attachment of the suction cup a different grasp pose is then attempted until a maximum number of attempts occurred. If no sweet peppers, grasp poses or peduncle poses were detected or the attachment attempts were exceeded the system moved into the move platform state. After all sweet pepper/s in a scene were attempted or harvested the system became free for movement to the next region of sweet peppers.

The ROS MoveIt! package \citep{moveit!} was used for executing the motion planning operations. Within the MoveIt! framework the Open Motion Planning Library \citep{ompl} implementation of RRT* \citep{rrtstar} was selected, along with the TRAC-IK \citep{Beeson2015} inverse kinematics (IK) solver. The distance optimisation setting of TRAC-IK was used to minimise the joint-space traversed through movement operations. We found the combination of RRT* and TRAC-IK produced consistent plans more often than alternative readily available options \citep{Beeson2015}. Collisions were handled within the motion planning framework which accounted for self-collisions of the robot platform in addition to custom vertical crow row boundaries to ensure the robot safely planned within each crop row.

Control of Harvey was implemented with readily available ROS control packages. A joint trajectory action server was used which accepted a 7DoF joint trajectory from the ROS Moveit! motion planner and split the joint trajectories to the Universal Robots ROS joint controller and a custom ROS trajectory controller for the lift axis.

\section{Perception and Planning}\label{sec:perception_and_planning}
\vspace{-1pt}
In this section we present our perception system for determining grasp and cutting poses through the segmentation of sweet peppers and peduncles. To do this we first capture a view of the sweet pepper with an RGB-D camera. We then segment the sweet pepper from its surrounding environment using colour information (Section \ref{sec:detection}). The pose of the peduncle is then estimated by using a deep peduncle segmentor (Section \ref{sec:peduncle_detection}) to perform per-pixel segmentation. Grasp poses are then selected using surface information from the 3D segmentation of the sweet pepper (Section \ref{sec:grasp_selection}). The grasp pose and peduncle pose are then used by the subsequent motion planning method (Section \ref{sec:motion_planning}) to perform the harvesting operation. 

The first stage of the perception pipeline involves two steps, firstly an image is captured from a long-range perspective in order to detect the initial location of a target sweet pepper. A target sweet pepper is selected based on distance to the origin of the robot arm workspace. The camera is then moved into a close-range perspective based on the position of the target sweet pepper but offset vertically in order to maximise the view of the peduncle. In this work the RGB-D sensor is an Intel\textregistered RealSense SR300 RGB-D camera and provides RGB colour and depth information used to create a colour point cloud of the scene.

\subsection{Sweet Pepper Segmentation}\label{sec:detection}

Sweet pepper segmentation is necessary to differentiate the sweet pepper from the background (leaves and stems). Per-pixel segmentation is necessary for the later stage of grasp pose selection. This task is challenging due to variation in crop colour and illumination as well as high levels of occlusion. An ad-hoc combination of features (local binary patterns, histogram of gradients, stacked auto-encoders and HSV) were used in our prior work~\citep{McCool:2016aa} to train a conditional random field that was capable of detecting both green and red sweet pepper.

In this work, we are only interested in segmenting red (ripe) sweet pepper and so we make use of a simpler computationally efficient method based purely on colour described in~\cite{Lehnert2016}.
A HSV colour segmentation algorithm is trained to detect ripe (in our case red) sweet peppers.  We convert the RGB information into the more consistent rotated hue, saturation, and value (rotated-HSV) colour space. The rotated-HSV colour space is chosen as its dependence upon intensity is assigned to a single dimension (V) rather than across all three RGB dimensions. The hue component is rotated by $90^\circ$ to avoid red values from lying on the border at $0^\circ$ and $360^\circ$. 

The distribution of red sweet pepper pixels is then modelled using a multivariate Gaussian. The likelihood that a pixel is from a red sweet pepper is then evaluated for each pixel.
\begin{equation}
    p\left( \vecx \mid \mean, \variance \right) = \left(2\pi\right)^{-\frac{1}{2}} \det\left|\variance \right|^{-\frac{1}{2}} \exp\left[ -\frac{1}{2} \left(\vecx - \mean \right)^T \variance^{-1} \left(\vecx - \mean \right) \right].
\end{equation}
The model parameters ($\mean$ and $\variance$) are learnt on a training set of manually annotated images; $\variance$ is assumed to be diagonal. This is a computationally efficient model to use as the log-likelihood reduces to
\begin{equation}
    \log\left[ p\left( \vecx \mid \mean, \variance \right)\right ] = -\frac{1}{2}\log\left(2\pi\right) -\frac{1}{2} \log \det \left|\variance \right| -\frac{1}{2} \left(\vecx - \mean \right)^T \variance^{-1} \left(\vecx - \mean \right) 
\end{equation}
and the first two terms, $-\frac{1}{2}\log\left(2\pi\right)$ and $-\frac{1}{2} \log \det \left|\variance \right|$, can be pre-computed.

\begin{figure}[tbh]
	\centering
				\includegraphics[width=\textwidth]{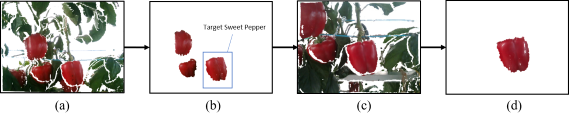}
		
	\caption{Diagram illustrating the steps performed to obtain a segmentation of a single target sweet pepper. (a) Firstly a Long-range perspective point cloud is captured (b) Then an initial sweet pepper segmentation is performed using colour, clustering and noise filtering to estimate a target sweet pepper which is closest to the workspace of the robot (c) The RGB-D camera is then moved to a close-range perspective centred on the target sweet pepper (d) A final sweet pepper segmentation is performed to produce a coloured point cloud of the sweet pepper used for grasp selection.}
	\label{fig:sweet_pepper_detection} 
\end{figure}

The output of the image segmentation is then used to segment the 3D points of sweet pepper from the colour point cloud of the scene. A resultant 3D point cloud containing only segmented sweet pepper points, is then clustered and filtered. A Euclidean clustering step is used to 
separate the points for multiple sweet peppers from the scene based on a minimum distance threshold. If multiple sweet peppers are in view this clustering step also determines the best candidate sweet pepper based on the cluster which has the largest number of segmented 3D points (most information available) and has the closest centroid to the end effector. Lastly, smoothing and outlier removal is performed on the points to filter noise from the segmentation step. The resulting output of this stage is the segmented point cloud of a single target sweet pepper. The stages of the sweet pepper segmentation method are illustrated in Fig~\ref{fig:sweet_pepper_detection}.

\subsection{Peduncle Segmentation}\label{sec:peduncle_detection}

It is highly desirable to be able to precisely segment and localise the peduncle before performing any crop cutting because retaining the proper peduncle maximises the storage life and market value of each crop. In addition, accurate peduncle segmentation can lead to higher success rates for crop detachment, which in turn yields more harvested crops. It is, however, a challenging task attributed by multiple factors; the presence of occlusions by leaves or other crops, varying lighting conditions in product environments, visually similar to other parts of the plant, and the natural variance in its shape (e.g., flat or highly curved).

Our previous work in \citet{Sa17_1} proposed a peduncle segmentation system to address these challenges based on hand-crafted colour and geometry (point feature histograms~\citep{Rusu2009-kl}) features used with a support vector machine (SVM), referred to as \textit{PFH-SVM}. Although this conventional approach is straight-forward and efficient in feature extraction and prediction, it suffers from two downsides. 
First, it is sensitive to variations in environmental conditions (e.g., varying lighting) which implies that the trained model can overfit to the environments where the dataset was collected and might not be generalisable.
Second, annotating the data is challenging and time-consuming as it requires the visualisation and selection of regions in a 3D viewer.

Inspired by the recent advances in computer vision, we consider the efficient deep learning method proposed by \citet{McCool17_1} to train a peduncle segmentation system using primarily the colour image.
This system has the potential to provide a highly accurate and efficient system which can be rapidly deployed to different environments, and tasks, due to the ease with which data can be collected and annotated. 
However, due to the importance of the 3D structure of the plant we employ a secondary filtering process.
This allows us to enforce structural constraints based on reasonable assumptions about the crop structure and is applied to both the \textit{PFH-SVM} and efficient deep learning approach for experimental comparison.
Below we describe the efficient deep learning approach and 3D filtering process.

\begin{figure*}[tbh]
	\centering
		\includegraphics[width=\textwidth]{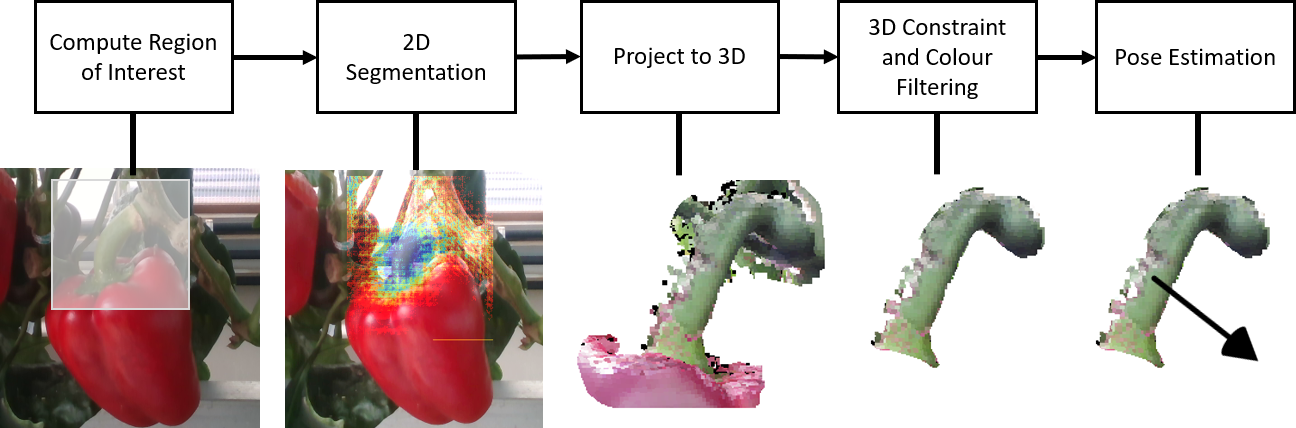}
	\vspace{-15pt}
	\caption{Steps for CNN peduncle localisation. Firstly a region of interest is computed given prior detected sweet pepper points (Compute Region of Interest). The ROI masked image is then passed into the CNN which returns detected peduncle points (2D Detection, blue and red denote high and low confidence regions respectively). The detected points are then projected to Euclidean space using the corresponding depth image from the RGD-D camera (Project to 3D). The projected points are then filtered using 3D constraints, colour information and cluster size (3D Constraint and Colour Filtering). The filtered peduncle points are then used to estimate a cutting pose aligned with the centroid of the points (Pose Estimation).   }
	\label{fig:pipe_line}
	\end{figure*}

\subsubsection{Deep Learnt System: lightweight agricultural DCNN}

\begin{figure*}[tbh]
	\begin{center}
		\begin{subfigure}[b]{0.32\textwidth}
			\includegraphics[width=\textwidth]{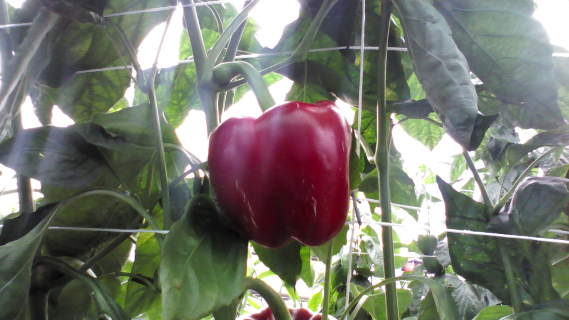}
			\caption{}
		\end{subfigure}
				\begin{subfigure}[b]{0.32\textwidth}
			\includegraphics[width=\textwidth]{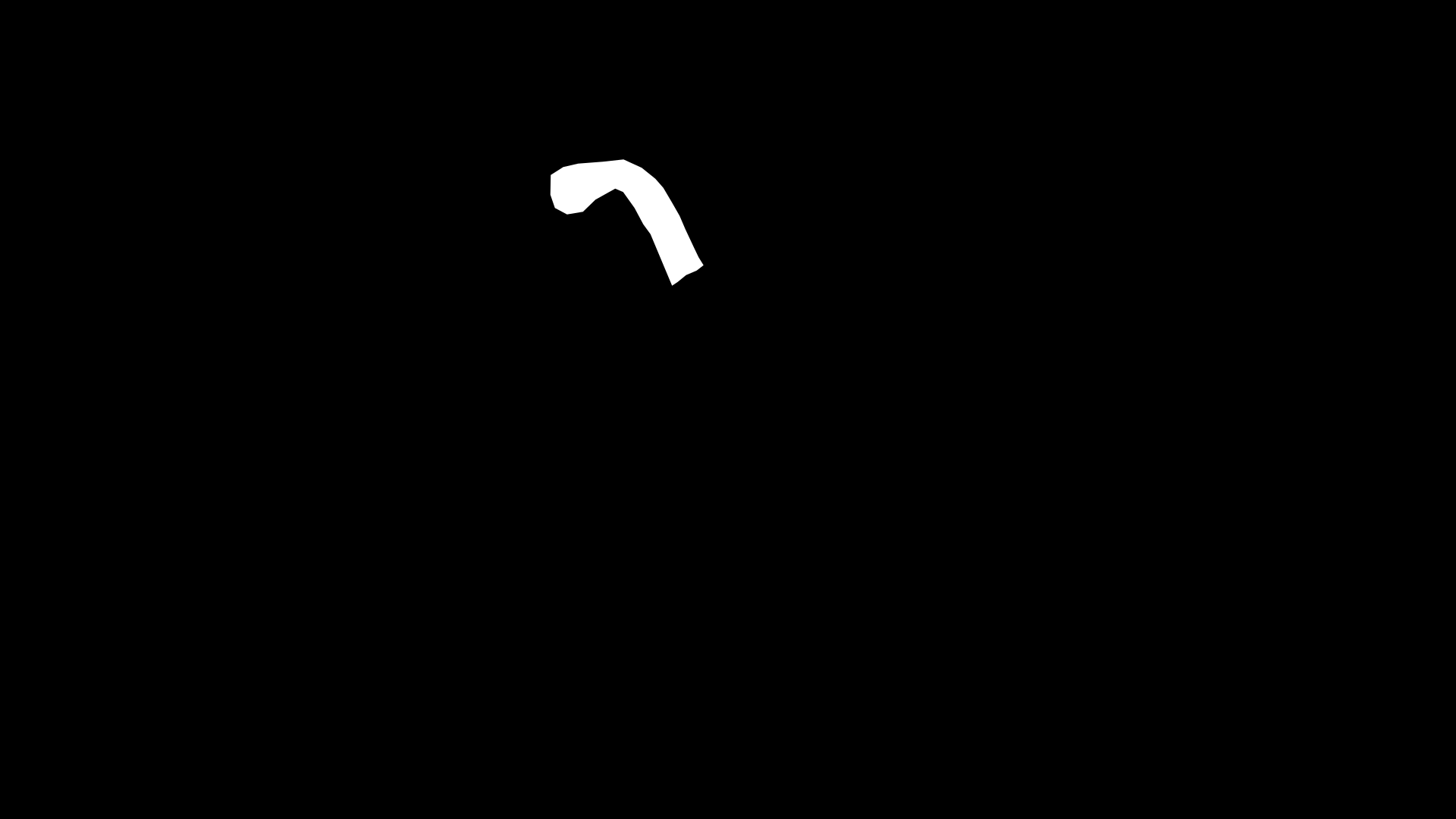}
			\caption{}
		\end{subfigure}
				\begin{subfigure}[b]{0.32\textwidth}
			\includegraphics[width=\textwidth]{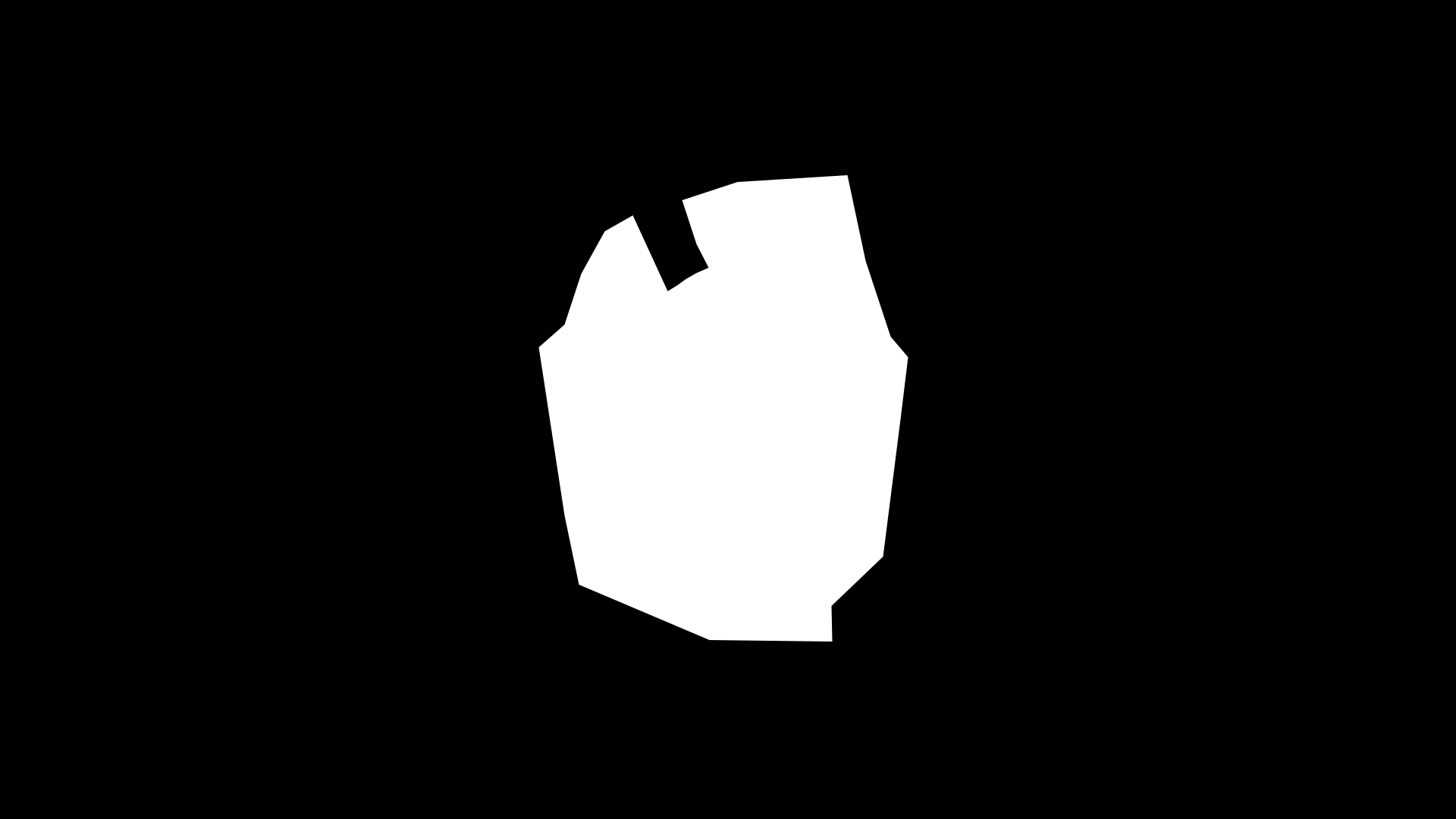}
			\caption{}
				\end{subfigure}	
\end{center}
	\caption{Example of manually annotated ground truth for peduncles. The original image is shown in (a) and the annotated peduncle (b) followed by the regions which do not represent the peduncle in (c).}
	\label{fig:cnn_annotations}
\end{figure*}

\citet{McCool17_1} proposed an approach for training deep convolutional neural networks (DCNNs) that allows for the tradeoff between complexity (e.g. memory size and speed) with accuracy while still training an effective model from limited data. We use the \textit{MiniInception} approach of \citet{McCool17_1} to train a lightweight DCNN for efficient peduncle segmentation.
When training a model like this it is normal to define the positive region, in this case the peduncle, and then consider everything else to be a negative example.
However, due to scene complexity this is not appropriate for our work as some parts of the scene may contain other peduncles, as can be seen in Fig~\ref{fig:cnn_annotations} (a).
As such, for each image we annotate the positive region, see Fig~\ref{fig:cnn_annotations} (b), as well as the negative region, see Fig~\ref{fig:cnn_annotations} (c).

\subsubsection{Region of Interest Selection and 3D Filtering} 

For a deployed system, the task of peduncle detection is performed once the sweet pepper has been detected.
Therefore we employ assumptions based on the structure of the plant improve the accuracy of the localisation.
We make use of two assumptions.
First, we can improve the efficiency of the two algorithms by pre-computing a 2D region of interest (RoI) so that only the region in the image above the detected sweet pepper is considered to contain the peduncle.
Similarly, 3D constraints such that the peduncle cannot be too distant from the detected sweet pepper are enforced using a 3D bounding box before finally declaring the position and pose of the peduncle.
An example of this process is given in Fig~\ref{fig:pipe_line}, this process is applied to both algorithms.

The 2D RoI is defined to contain the region within the image above the detected sweet pepper. 
Given a bounding box of the detected sweet pepper of height $h_b$ and width $w_b$ and central position $\left( c_{x}, c_{y} \right)$, the region of interest for the peduncle is defined to have the same width, $w_b$, and height, $h_b$. 
The central location of the peduncle RoI is then given shifting up by half the height of the sweet pepper bounding box, $\left( c_{x}, c_{y} + h_{b}/2 \right)$.

In order to improve the detection of peduncle points a corresponding depth map is used to filter the points using Euclidean constraints.
The filtering steps used are as follows: 
\begin{enumerate}[topsep=0pt,itemsep=-1ex,partopsep=1ex,parsep=1ex]
    \item threshold the classification scores,
    \item project thresholded scores and corresponding depth points to a point cloud,
    \item detect and delete capsicum points using HSV detector,
    \item delete points outside a 3D bounding box and
    \item perform Euclidean clustering on the 3D point cloud and select the largest cluster.
\end{enumerate}

\begin{figure}[tbh]
\centering
\includegraphics[height=6.5cm]{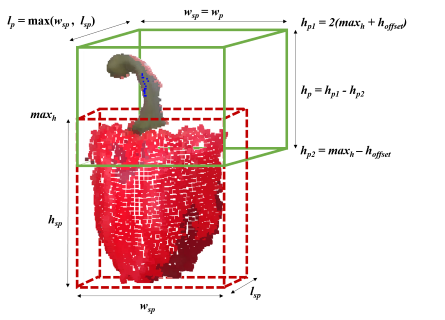}
\vspace{-10pt}
	\caption{Illustration of how a 3D peduncle constraint bounding box is calculated based on the 3D Bounding Box (BB) from the detected sweet pepper. Where $w_{sp}, l_{sp}, h_{sp}$ correspond to the width, length and height of the sweet pepper point cloud and $w_p, l_p, h_p$ corresponds to width, length and height of the calculated peduncle 3D BB. $max_h$ corresponds to the maximum height of the sweet pepper and $h_{offset}$ is a offset parameter for defining the upper and lower heights of the peduncle 3D BB.}
	\label{fig:bbox_diagram}
\vspace{-5pt}
\end{figure}

The 3D Bounding Box (BB) is used to delete peduncle outliers using the maximum and minimum euclidean points from the detected sweet pepper points. 
The definitions of the width, length and height of the BB are given in Fig~\ref{fig:bbox_diagram} and shows that the length, $l_p$ of the peduncle BB is given by the max of either the width, $w_{sp}$, or length, $l_{sp}$ of the sweet pepper BB.
The reason for selecting the max of width or length is due to the fact that the depth points of the sweet pepper are measured from one side (view) only and it is assumed that the sweet pepper is symmetrical about this axis. 
Therefore, the largest measure, width or length, gives the maximum BB of the sweet pepper in those axes. Furthermore, the height $h_p$ of the peduncle BB is defined by the max height of the sweet pepper, $max_h$, and a predefined height offset parameter, $h_{offset}$. For this work we defined the height offset parameter as \unit{50}{mm} which is the average length ($N=25$) of a peduncle for the varieties in the field tests.

\subsection{Grasp Selection} \label{sec:grasp_selection}

Grasp poses for each sweet pepper are calculated using the segmented 3D point cloud of a sweet pepper. This work uses the method presented in \citep{Lehnert2017} for selecting grasp poses. A summary of this method is presented in this section.

The grasp selection method finds multiple grasp poses from point cloud data by computing the surface normals over the points using a fixed patch size. These surface normals are used as initial candidate grasp poses and subsequently ranked based on a utility function. The utility function is the weighted average of three normalised scores $S_{i1}, S_{i2} \text{ and } S_{i3}$ based on the surface curvature, distance to the point cloud boundary and angle with respect to the horizontal world axis, respectively, where $i$ is the current candidate pose. This utility function favours grasp poses that are close to the centre of the sweet pepper, on planar surfaces, are aligned with the horizontal world axis and away from discontinuities caused by occlusion. 
The \textit{utility}, $U_i$ of the grasp pose $i$ is calculated according to
\begin{equation}
U_i=\sum_{j=1}^{3} W_j \, S_{ij}, \text{ given } \sum^{3}_{j=1} W_j  =1
\end{equation}
where $S_{ij}$ is the normalised \textit{score} of the grasp pose $i$ and $W_j$ are \textit{weighting coefficients} that describe the importance of each score.

\begin{figure}[tbh]
		\centering
		
	\begin{subfigure}[b]{\columnwidth}\centering
						
		\includegraphics[height=6.25cm]{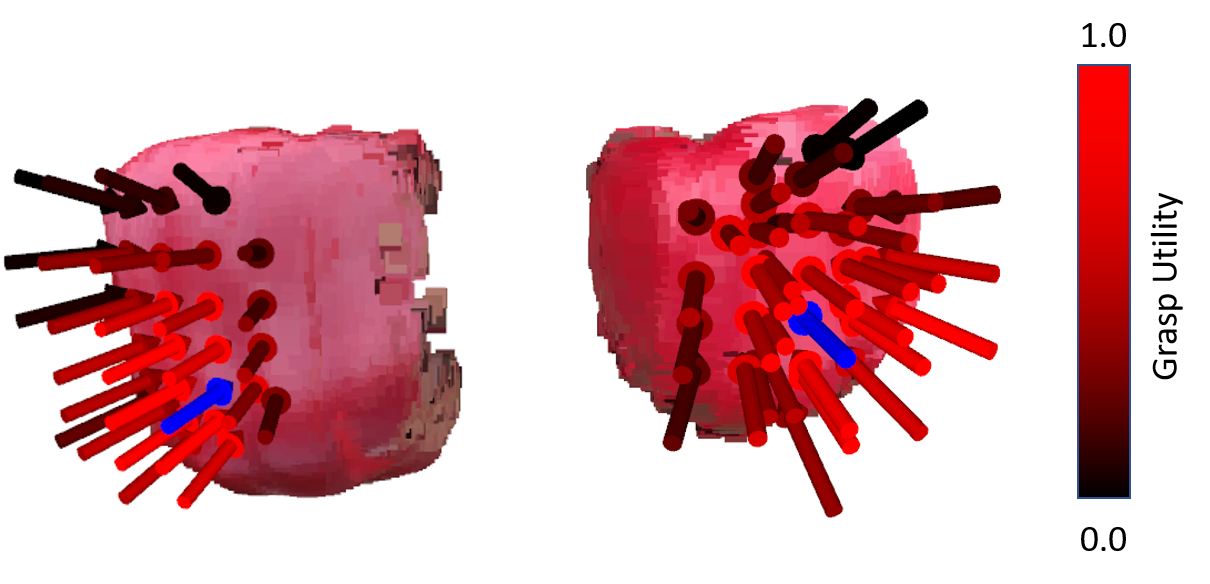}
	\end{subfigure}
	\vspace{-15pt}
	\caption{Computed grasp poses. The utility of each grasp pose is shown as the gradient from red to black, where black is the lowest utility. The blue arrow indicates the best pose based on its utility.}
	\label{fig:grasp_detection}
	\vspace{-15pt}
\end{figure}

An example of the grasp selection method applied to real sweet pepper point clouds is shown in Fig~\ref{fig:grasp_detection} where the utility of each grasp pose is represented as the gradient from red to black, where black has the lowest utility and the blue pose indicates the grasp pose with the highest utility. An added advantage of this grasping method is that if the grasp pose with highest utility is unsuccessful the next candidate pose can be used.

\subsection{Attachment and Detachment Motion Planning} \label{sec:motion_planning}

\begin{figure}[tbh]
	\centering
	\includegraphics[height=7.5cm]{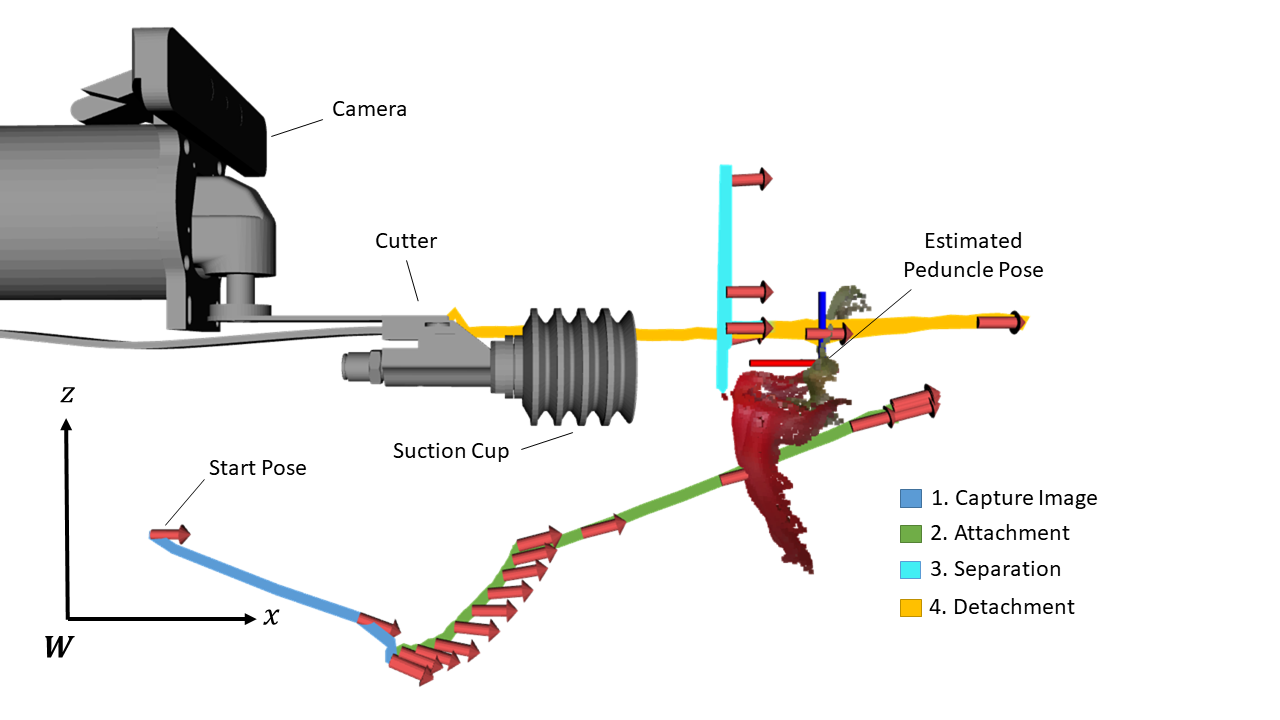}
	\caption{End effector trajectory of a single harvesting trial for the estimated pose of a sweet pepper. The position of the end effector is indicated by the coloured lines whereas the orientation is represented by the red arrows. The trajectory begins (start pose) with the attachment stage indicated by the yellow line, transitioning into the separation stage as the blue line corresponding to a vertical motion and finishing with the detachment stage illustrated as the green line.}
	\label{fig:harvest_traj}
\end{figure}

Motion planning is performed sequentially for attachment and then detachment.
The attachment trajectory starts at a fixed offset from the sweet pepper determined by the close range image capture location and moves from the close range pose to a pre-grasp pose which has a fixed offset along the approach direction. The trajectory then makes a linear movement towards the selected grasp pose causing the suction cup to attach to the sweet pepper. This motion computed by the planner can be seen within Fig~\ref{fig:harvest_traj}, depicted by the green line. A fixed offset is also applied to the final grasp pose of the attachment trajectory in order to ensure the suction cup makes a seal (indicated by the green line drawn within the sweet pepper).  

Once the attachment trajectory has been executed attaching the suction cup, the end effector is moved vertically along the $\bf{z}$ axis of the world frame from the final grasp pose in order to decouple the suction cup from the cutting tool (as described in  Section \ref{sec:harvest_tool}). 

The final step for the motion planner is to compute a cutting trajectory. It was found that using a cutting trajectory that is aligned with the $\bf{x}$ axis of the world frame performed better than when aligned with the orientation of the peduncle. This method worked as estimating the orientation of the peduncle was sensitive to the number of 3D points detected. The orientation of the cutting tool is kept level during the trajectory. An example detachment trajectory is shown as the yellow line in Fig~\ref{fig:harvest_traj} and includes both the inwards and outwards motion of the cutting tool which has a fixed orientated (red arrows) along the world x axis.

The resulting end effector trajectory for the attachment, separation and detachment stages for one harvesting cycle is shown in Fig~\ref{fig:harvest_traj}. This figure depicts the trajectory of the end effector relative to the estimated pose of the target sweet pepper, highlighting the three steps: capture image, attachment, separation and detachment of the harvesting process. 

As discussed in Section~\ref{sec:cropping_environment}, operating in a protected cropping environment with relatively planar row structure simplifies the planning an manipulation tasks so that we do not need to perform complex obstacle avoidance to move between the branches of  a 3D plant structure. Our motion planner takes into account self collisions with the robot arm and base platform, and assumes a simple planar obstacle closely behind the crop row to reduce the probability of collisions with the plant. Trajectories that interact with the plant are generally ``in and out" motions that reduce the chance of collisions.

\section{Experiments and Results}
\label{sec:exp_results}

\begin{figure}[bh!]
	\centering
	\begin{subfigure}[b]{0.6\textwidth}\centering
		\includegraphics[height=5.5cm]{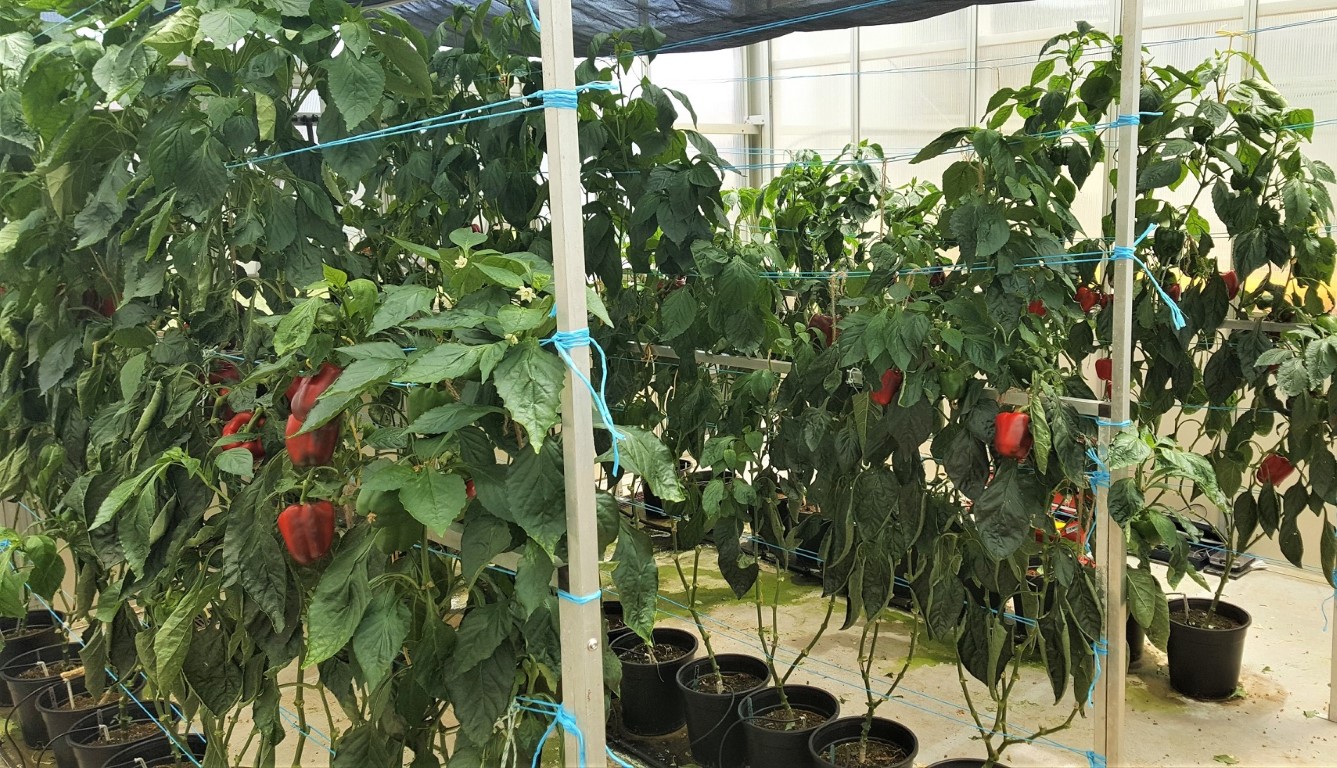}
		\caption{}
		\label{fig:crop_environment}
	\end{subfigure}
	\begin{subfigure}[b]{0.3\textwidth}\centering
		\includegraphics[height=5.5cm]{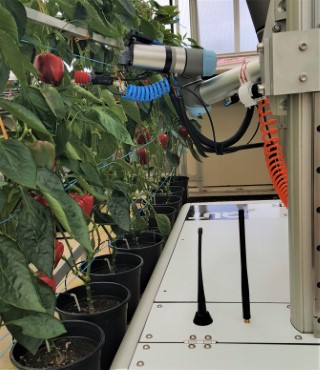}
		\caption{}
		\label{fig:platform_in_field}
	\end{subfigure}
						\caption{ Setup for field experiments in protected cropping system. (a) An image of the protected cropping environment used within the field experiments. The cropping environment includes a horizontal trellis system that was replicated from real protected cropping systems in Queensland. (b) image of the platform within the protected cropping environment, illustrating the workspace of the robotic arm and harvesting tool.}
	\label{fig:harvey_platform_exp}
	\vspace{-5pt}
\end{figure}

A field trial was conducted within a protected cropping research facility in Cleveland, Queensland (Australia) over a 5 day period. Overall the robot platform has been tested on a total of 68 sweet peppers in a real protected cropping system. Within this work two different sweet pepper cultivars were trialled, Mercuno and Ducati. 

Three experiments are presented in the following section aimed at validating the perception system and overall harvesting method. First, we present the accuracy of the sweet pepper segmentation system, to demonstrate that we can accurately segment out the ripe (red) sweet pepper of interest from the background. Second, we present experimental results for the peduncle segmentation system, comparing the performance of the deep learning method to a previous hand-crafted 3D feature method. Finally, we present a field experiment of the full harvesting platform in a real protected cropping system, demonstrating the harvesting performance of the final integrated system.

\subsection{Experimental Setup} \label{sec:experiment_setup}

Three crop rows were selected for the experimental work including a total of 68 sweets peppers. Each sweet pepper out of the total present within the crop rows were included in the experiment. An image of the crop rows within the protected cropping system for this experimental work is shown in Fig~\ref{fig:crop_environment} and example images of the platform within the environment are shown in Fig~\ref{fig:harvey_platform_exp}b.

\begin{figure}[tbh!]
	\centering
	\begin{subfigure}[b]{\textwidth}\centering
		\includegraphics[width=0.3\textwidth]{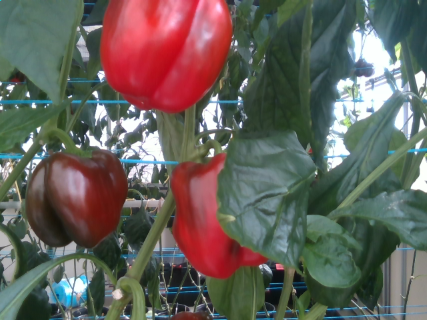}
		\includegraphics[width=0.3\textwidth]{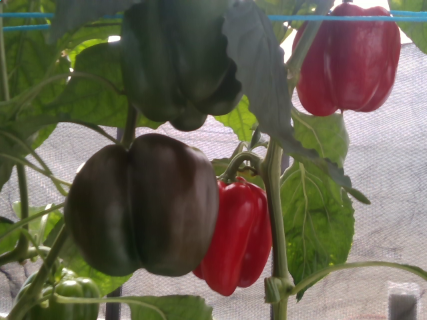}
		\includegraphics[width=0.3\textwidth]{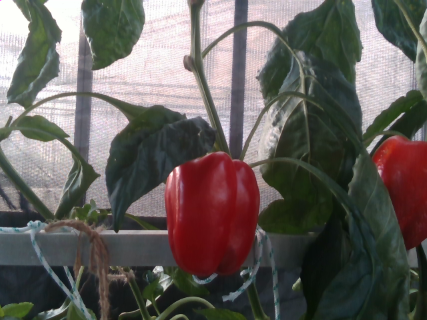}
		\caption{ Unmodified }
		\label{fig:modified_sweet_pepper:a}
	\end{subfigure}
	
	\begin{subfigure}[b]{\textwidth}\centering
		\includegraphics[width=0.3\textwidth]{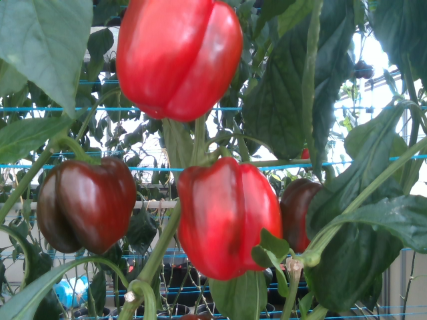}
		\includegraphics[width=0.3\textwidth]{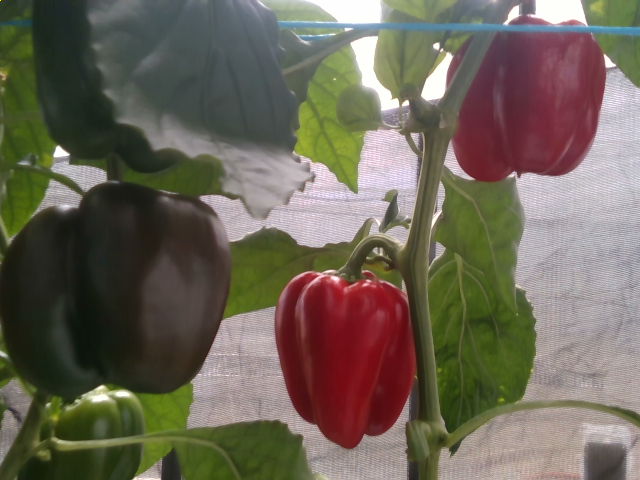}
		\includegraphics[width=0.3\textwidth]{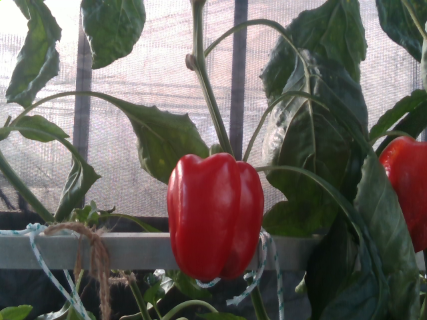}
		\caption{ Modified }
		\label{fig:modified_sweet_pepper:b}
	\end{subfigure}

	\caption{(a) Unmodified versus (b) modified sweet pepper, where leaves have been removed. (c) Unmodified versus (d) modified sweet pepper (no. 6) where leaves have been removed and the pose of the sweet pepper has been adjusted. For this case, the sweet pepper was adjusted to be in front of the trellis string and the main stem of the plant. }
	\label{fig:modified_sweet_peppers}
\end{figure}

The methodology for the field trial is as follows. The robotic platform was positioned at the beginning of a crop row. The robot was then commanded to perform a single harvest cycle. If the robot failed to detach the sweet pepper, the robot arm moved back to its start position and the attempt was retried. If obstructions or occlusions caused multiple failed attempts and that any further attempts would likely continue to fail the scene was modified by either removing leaves or by adjusting the position of the sweet pepper. Fig~\ref{fig:modified_sweet_peppers} shows examples of how the modifications to the experiment were performed. The robot was then commanded to repeat the attempt from the same starting position. The results of each attempt were recorded and details were noted such as if a modification was performed, whether a failure occurred, what was the cause of the failure if any and if damage to the sweet pepper or plant occurred.

Once no more sweet peppers could be detected from the robots current position the platform was moved forward via remote control by \unit[0.5]{m} (approximately half of the width of the cameras field of view). This distance was selected as it had sufficient overlap to detect sweet peppers from different perspectives if occluded in the previous view.

Each attempt was broken down into four parts where the measurement for success was:
\vspace{-0.3cm}
\begin{itemize}
	\setlength\itemsep{0.5pt}
    \item Sweet Pepper Detection---a sweet pepper was detected and grasp pose was found;
    \item Peduncle Detection---a peduncle was detected for the targeted sweet pepper;
    \item Attachment---the suction cup attached to the sweet pepper sufficient to maintain a grasp;
    \item Harvesting ---the peduncle was cut and a sweet pepper was successfully placed into a storage bin.
\end{itemize}
Additional notes were also recorded during the experiment on whether any damage to the sweet pepper or plant was caused during the harvesting process and categorised as:
\vspace{-0.3cm}
\begin{itemize}
	\setlength\itemsep{0.5pt}
	\item Major or minor damage to the sweet pepper;
	\item Major or minor damage to the plant or stem.
\end{itemize}

\begin{figure}[tbh!]
	\centering
	\begin{subfigure}[b]{0.45\columnwidth}\centering
		\includegraphics[height=5.5cm]{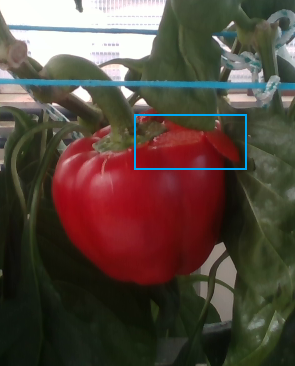}
		\caption{ }
		\label{fig:sweet_pepper_minor_dmg}
	\end{subfigure}
	\begin{subfigure}[b]{0.45\columnwidth}\centering
		\includegraphics[height=5.5cm]{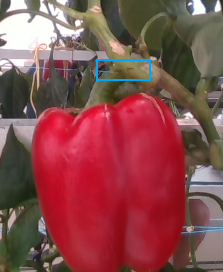}
		\caption{ }
		\label{fig:example_stem_minor_dmg}
	\end{subfigure}
	\caption{Example sweet pepper and plant damage. (a) Sweet pepper with damage highlighted in blue from a previous attempt. (b) Example plant damage highlighted in blue where the cutting blade was placed incorrectly. It can be seen that the peduncle of this sweet pepper is short and behind the plant stem making it difficult for peduncle detection.}
	\label{fig:example_sweet_pepper_dmg}
\end{figure}

The parameters for the sweet pepper segmentation, grasp selection, peduncle segmentation and attachment subsystems described in previous sections are given in Table~\ref{tbl:harvest_parameters} and were determined empirically through testing of the robotic harvesting system. Fr this work HSV model and threshold were the same as that used in previous work~\citep{Lehnert2017}. 

\begingroup
\newcommand{\e}{\ensuremath{ \mbox{\scriptsize{E}} }}
\renewcommand{\arraystretch}{1.5}
\begin{table}[tbh!]
	\centering
	\caption{Parameters for harvesting experiment.}
	\label{tbl:harvest_parameters}
	\begin{tabular}{clc}
		\Xhline{2.5\arrayrulewidth}
		Subsystem         & Parameter                              &                                    Value                                    \\ \Xhline{2.5\arrayrulewidth}
		Sweet Pepper Segmentation & HSV model parameters                   & \makecell{$\mean = [180, 1.0, 0.39]$, \\  $\variance = [255, 0.13, 0.017]$} \\
		& Point cluster size [min, max]          &                            [$1\e{3}$, $25\e{4}$]                            \\
		& Down sample radius                     &                               \unit[0.002]{m}                               \\ \Xhline{1\arrayrulewidth}
		Grasp Selection      & Curvature, boundary, rotational weight &                               [0.2, 0.5, 0.3]                               \\
		& Angle threshold                        &                                $\pi/4$ rads                                 \\
		& Surface normal patch size              &                               \unit[0.02]{m}                                \\ \Xhline{1\arrayrulewidth}
		Peduncle Segmentation   & Point cluster size [min, max]          &                               $[50, 25\e{4}]$                               \\
		& Down sample radius                     &                               \unit[0.002]{m}                               \\
		& Height offset constraint ($h_{offset}$)&                               \unit[0.05]{m}                                \\ \Xhline{1\arrayrulewidth}
		Attachment         & Max no. attempts                       &                                 5                                           \\ \Xhline{2.5\arrayrulewidth}
	\end{tabular}
\end{table}
\endgroup

\FloatBarrier
\subsection{Experiment 1: Sweet Pepper Segmentation}\label{sec:experiment_1}

Accurate segmentation of sweet pepper is a precursor for automated harvesting. In Section~\ref{sec:detection} we presented a method, based purely on colour, to pixel-wise detect ripe (red) sweet peppers. We evaluate the performance of this system by comparing to prior work~\citep{McCool:2016aa} on sweet pepper detection and quantitatively evaluate the performance using the area under the curve (AUC) of the precision-recall curve. Precision (P) and recall (R) are given by,
\begin{equation} \label{eq:precision_recall}
  P = \frac{T_{p}}{T_{p}+F_{p}}  \quad \text{and} \quad R = \frac{T_{p}}{T_{p}+F_{n}},
\end{equation}
where $T_p$ is the number of true positives (\textit{correct detections}), $F_p$ is the number of false positives (\textit{false alarms}), and $F_n$ is the number of false negatives (\textit{mis-detections}). Ideally, precision and recall should be 1 as this means there are no false positives and no false negatives.

For evaluation, we manually annotated a set of red sweet pepper images and divided this into a training and test set using a $2:1$ split. The training set consists of 20 images and the test set consists of 10 images. Using this, we trained both the CRF-based~\citep{McCool:2016aa} and our proposed colour-based approaches. 

\begin{figure}[tbh]
	\centering
	\includegraphics[height=7cm]{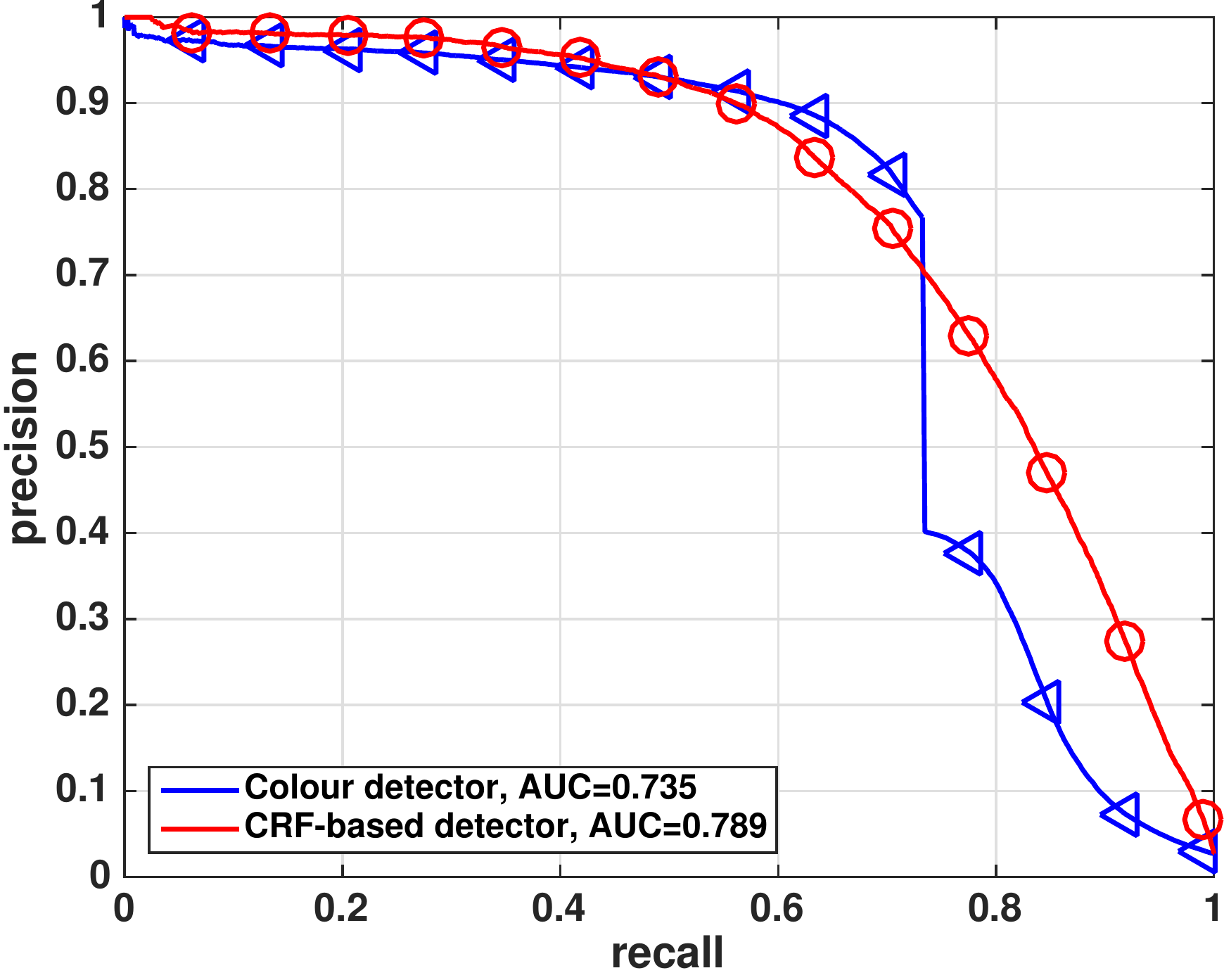}
		\caption{Precision-recall curve for detecting red sweet pepper using the CRF-based approach (red) and the colour-based approach (blue).}
	\label{fig:detector_AUC}	
	\end{figure}

The results in Fig~\ref{fig:detector_AUC} show that for the majority of the precision-recall curve the CRF-based and colour-based approaches have similar performance. It can be seen that once the recall exceeds 0.7, the performance of the colour-based approach drops considerably. By comparison, the performance of the CRF-based approach degrades consistently. This leads to the CRF-based approach achieving an AUC of 0.789 compared to the colour-based approach which achieves an AUC of 0.735. We attribute this performance difference to the reliance of a single features, for the colour-based approach, compared to the CRF-based approach which makes use of 4 features as well as taking into account the neighbouring information for the inference.

From these results we conclude that for the task of detecting ripe (red) sweet pepper it is sufficient to use the computationally efficient colour-based detector. However, if we were to change the task to also pick green sweet pepper then the CRF-based detector should be used; future work will explore the potential of such a system. 

\FloatBarrier
\subsection{Experiment 2: Peduncle Segmentation}\label{sec:experiment_2}

We present results for the two algorithms, \textit{PFH-SVM} and \textit{MiniInception}, executed on Harvey for detecting and segmenting peduncles. A small form GeForce 1070 was used for inference of the \textit{MiniInception} model. The system was deployed in a glasshouse facility in Cleveland (Australia) and consisted of two cultivar \textit{Ducati} and \textit{Mercuno}.
To train the \textit{MiniInception} approach, 41 annotated images were used. These images came from two sites, 20 images were obtained from the same site in Cleveland several weeks prior to deploying the robot which included a different set of crop on the plant and 21 were obtained from another site in Giru, North Queensland.

\subsubsection{Segmentation Performance}

\begin{figure}[!htb]
    \centering
    \begin{subfigure}[b]{0.49\textwidth}\centering
		\includegraphics[trim={13cm 0cm 13cm 0cm},clip, width=\columnwidth]{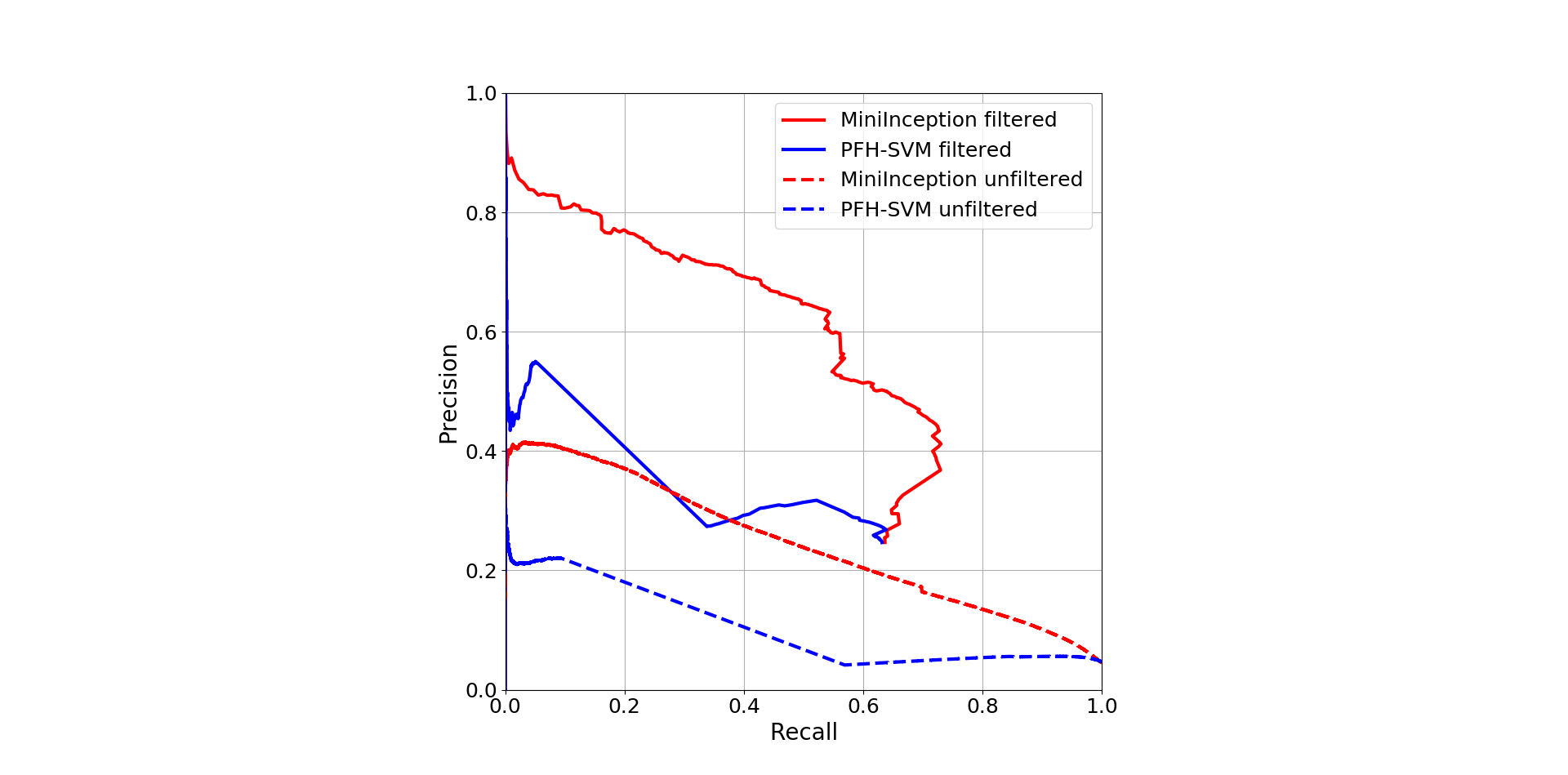}
		\caption{}
	\end{subfigure}
	\begin{subfigure}[b]{0.49\textwidth}\centering
		\includegraphics[trim={13cm 0cm 13cm 0cm},clip, width=\columnwidth]{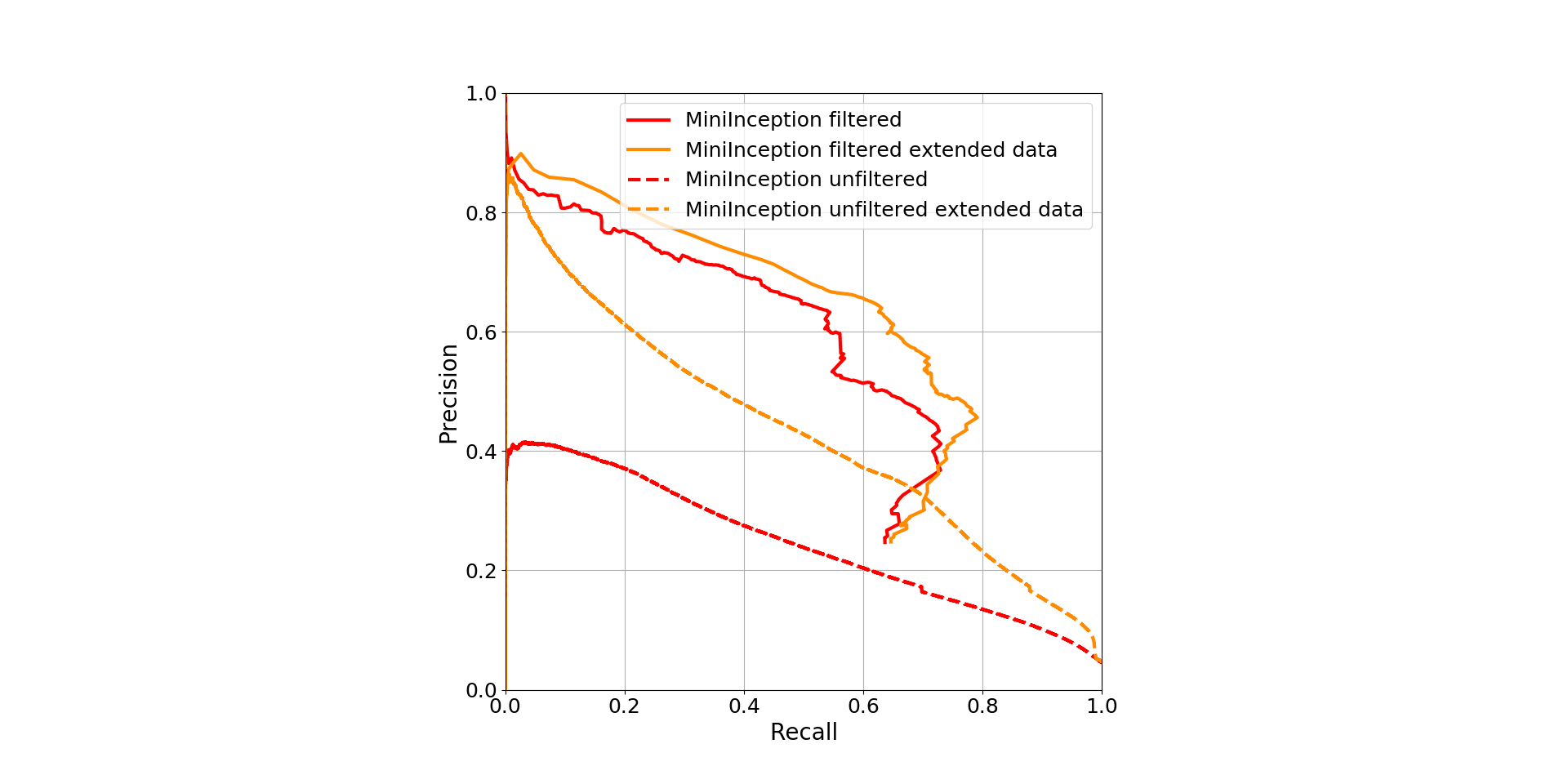}
		\caption{}
	\end{subfigure}	  
    \caption{(a) Precision recall results for the \textit{MiniInception} (DeepNet) and \textit{PFH-SVM} (PFH) algorithms before and after the filtering step and (b) A comparison between using \textit{MiniInception} with and without extended training data.}
    \label{fig:pr_dcnn_vs_trad}
\end{figure}

The performance of the two algorithms, \textit{PFH-SVM} and \textit{MiniInception}, is summarised in Fig~\ref{fig:pr_dcnn_vs_trad}a.
It can be seen that the performance of the \textit{MiniInception} model is consistently superior to that of the \textit{PFH-SVM} approach.
However, both approaches have relatively low performance with $F_1$ scores of 0.313 and 0.132 for the \textit{MiniInception} and \textit{PFH-SVM} systems respectively.

On average the execution time of the two algorithms is similar with the \textit{MiniInception} approach executing an average of 1704 points per second while the \textit{PFH-SVM} approach executes at an average of 1248 points per second. This measurement is reported as the two methods receive a different number of points (3D points vs 2D pixels) for the same data.

Introducing the filtering step described in section \ref{sec:peduncle_detection} provides a considerable improvement in performance for both algorithms.
The $F_1$ for the \textit{MiniInception} and \textit{PFH-SVM} systems improve to 0.564 and 0.302 respectively.

\begin{figure}[tbh]
    \centering
	\begin{subfigure}[b]{0.33\columnwidth}\centering
		\includegraphics[height=5cm]{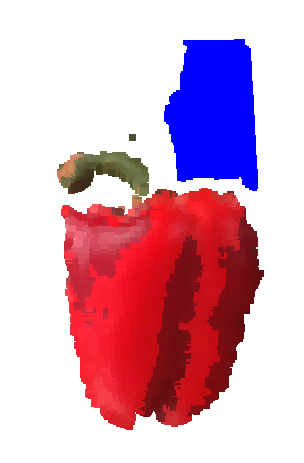}
		\caption{}
	\end{subfigure}
	\begin{subfigure}[b]{0.33\columnwidth}\centering
		\includegraphics[height=5cm]{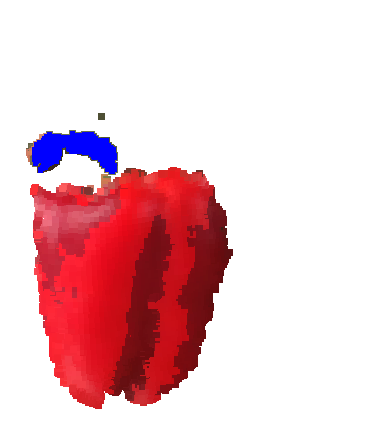}
		\caption{}
	\end{subfigure}	  
	\caption{Example behaviour of the post filtering results where blue represents the classified peduncle points. (a) Example segmentation at low precision (low threshold) where the maximum cluster size has selected a cluster that is a separate leaf or plant and fits within the 3D constraints. (b) Same example with higher precision---leaves and plant are thresholded out resulting in the cluster of the peduncle been selected.}
    \label{fig:pr_curve_recall_problem}
\end{figure}

For both algorithms, introducing the filtering step leads to odd behaviour in the precision-recall curve.
This is expected because we are altering the threshold on an algorithm, either \textit{MiniInception} or \textit{PFH-SVM}, whose performance is dependent on another step greatly impacting its final result.
An example of this is illustrated in Fig~\ref{fig:pr_curve_recall_problem} where introducing the filtering step at different thresholds leads to different points of the point cloud being considered as peduncles, and other points being suppressed. At low precision with high recall (low threshold) based on the assumption of selecting the maximum cluster size, a cluster that is a separate leaf or plant which fits within 3D constraints may be selected (see Fig~\ref{fig:pr_curve_recall_problem}a). Once the precision becomes higher, leaves and plant are thresholded out and the maximum cluster assumption becomes valid, resulting in the cluster of the peduncle been selected (see Fig~\ref{fig:pr_curve_recall_problem}b). Therefore in order for a peduncle cluster to be selected a minimum level of precision (threshold) is required.

\subsubsection{Qualitative Results}

\begin{figure}[tbh]
	\centering
	\includegraphics[width=\textwidth]{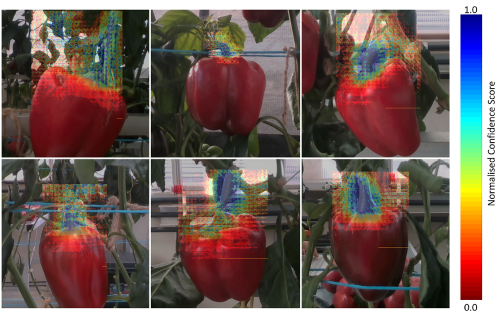}	
	\caption{Example outputs from the MiniInception model. Images are overlaid with normalised confidence scores from the CNN. It can be seen that majority of confidence scores are high on peduncle pixels. Some high confidence scores can also be seen on the stem and leaves of the plant but are mostly sparse. Majority of false positives from stem and leaf segmentations can be filtered out using further Euclidean clustering and constraints on the resulting point cloud of all segmented points}
	\label{fig:CNN_image_results}
\end{figure}

Qualitative results for the \textit{MiniInception} segmentation algorithm are presented in Fig~\ref{fig:CNN_image_results}.
From these results it can be seen that the deep network approach provides consistent results across multiple poses.
Also, it can be seen that the regions with high scores surround the peduncle region.
We believe this, in part, explains the poor precision-recall curve for the \textit{MiniInception} algorithm as these points will be considered as false positives and greatly reduce the precision value.
This is despite their proximity to the peduncle.
This also explains the considerable gain achieved by introducing the filtering step as many of these points will correspond to background regions and be discarded.

\begin{figure*}[tbh]
	\centering
	\begin{subfigure}[b]{0.19\textwidth}\centering
		\includegraphics[trim={4cm 1cm 4cm 1cm},width=0.95\textwidth]{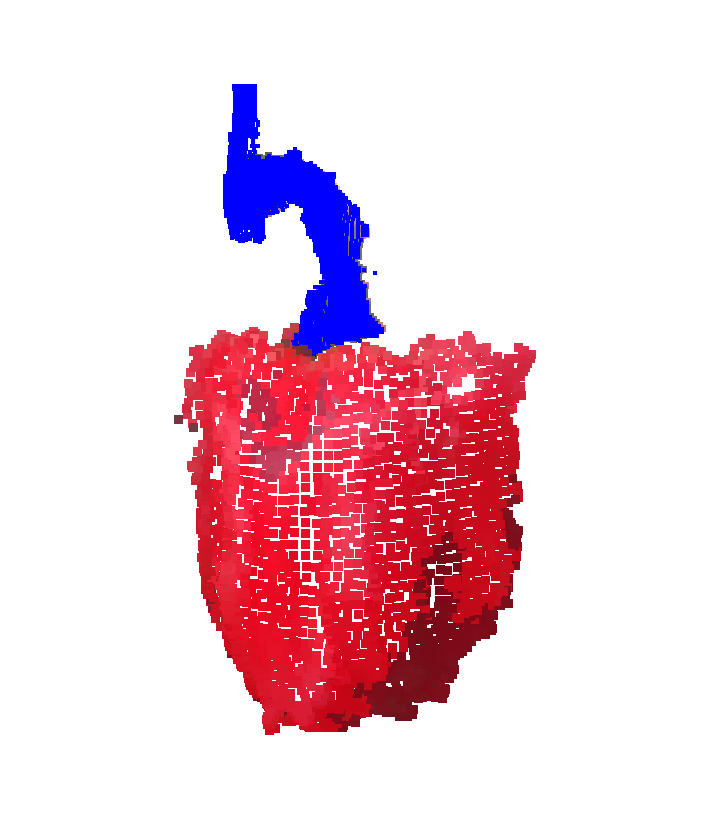}	
	\end{subfigure}
	\begin{subfigure}[b]{0.19\textwidth}\centering
		\includegraphics[trim={4cm 1cm 4cm 1cm},width=0.95\textwidth]{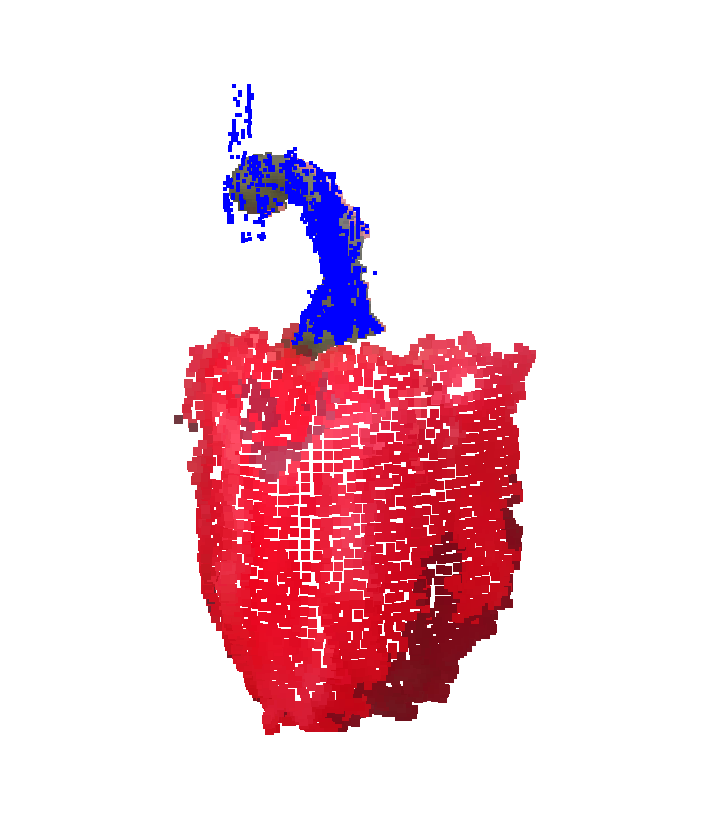} 
	\end{subfigure}	
	\begin{subfigure}[b]{0.19\textwidth}\centering
		\includegraphics[trim={4cm 1cm 4cm 1cm},width=0.95\textwidth]{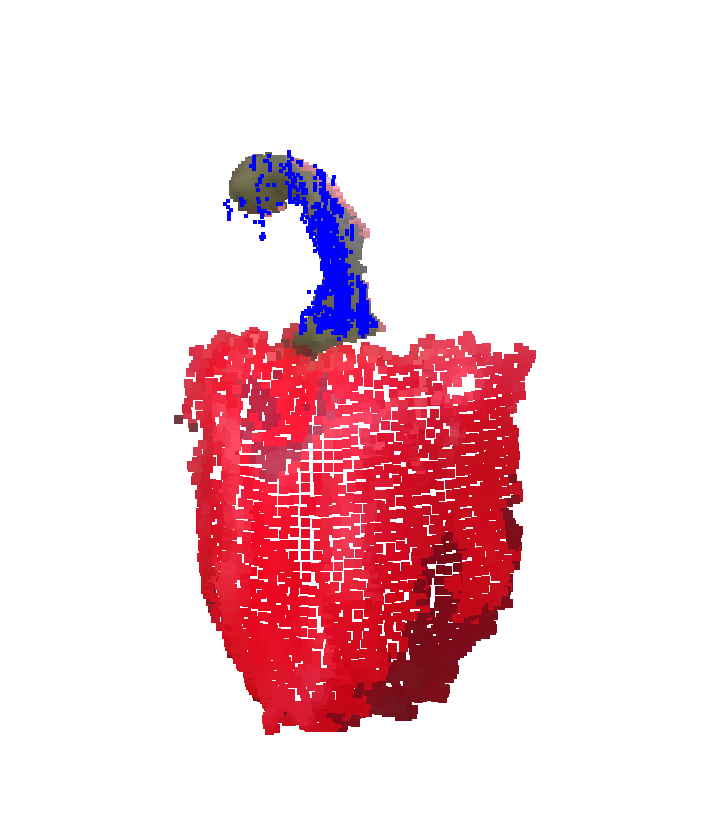} 	
	\end{subfigure}
	\begin{subfigure}[b]{0.19\textwidth}\centering
		\includegraphics[trim={4cm 1cm 4cm 1cm},width=0.95\textwidth]{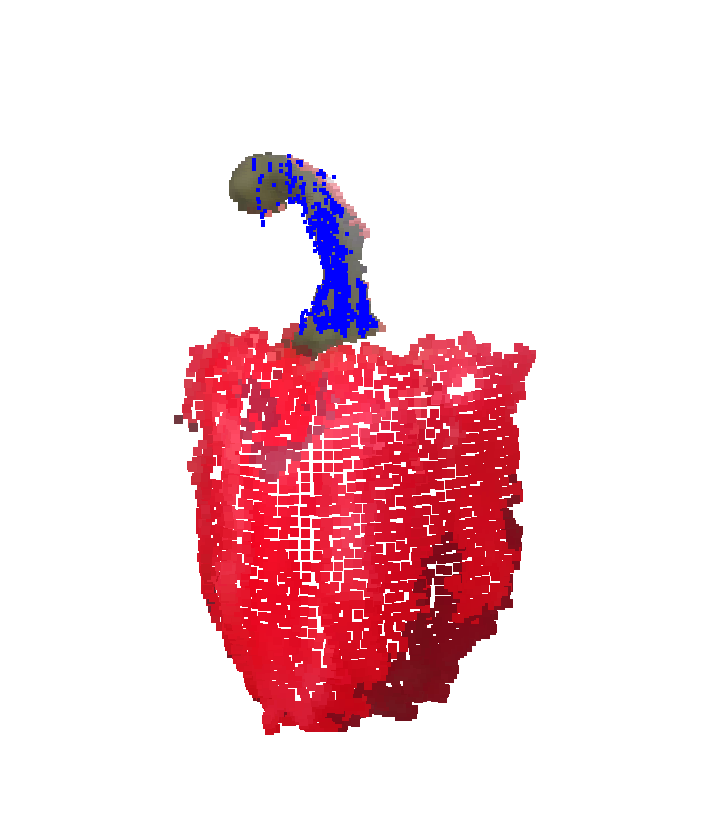}
	\end{subfigure}	
	\begin{subfigure}[b]{0.19\textwidth}\centering
		\includegraphics[trim={4cm 1cm 4cm 1cm},width=0.95\textwidth]{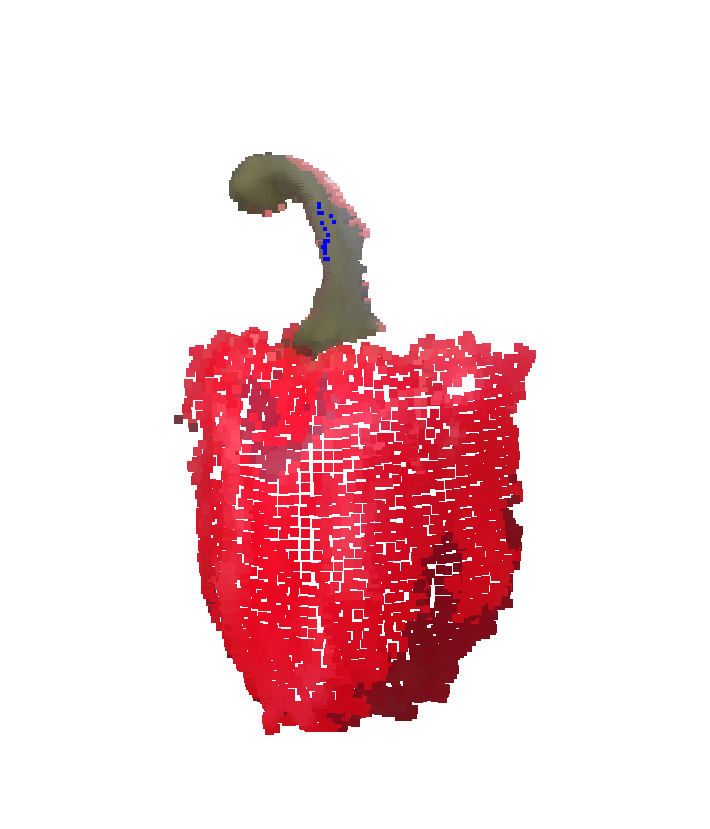}
	\end{subfigure}
	\caption{Example peduncle segmentation after filtering from CNN responses with varying threshold values and projected to 3D points using a corresponding depth map.
	The threshold value increases from left to right. The segmented peduncle points are highlighted in blue.}
	\label{fig:CNN_and_pfh_filter_results}
\end{figure*}

In Fig~\ref{fig:CNN_and_pfh_filter_results} we present example results of the entire procedure for the \textit{MiniInception} algorithm, with filtering, at varying thresholds.
It can be seen that as the threshold is increased the erroneous points, such as those belonging to the stem, are removed.
Even at higher threshold values a large number of points on the peduncle of the fruit are chosen.

\subsubsection{Extended Training Data}

One of the advantages of the \textit{MiniInception} approach is that it is much easier to annotate training data than the \textit{PFH-SVM} approach.
To determine if this can be beneficial we extended the training set of \textit{MiniInception} with an extra 33 images to a total of 74 annotated images. The extra images came from the Cleveland test site after performing the final harvesting experiment. Another purpose of the additional images is to investigate the potential improvement in performance when using additional domain specific images. This system is referred to as \textit{MiniInception-Extended}.

In Fig~\ref{fig:pr_dcnn_vs_trad}b it can be seen that the \textit{MiniInception} approach benefits considerably by increasing the training set size.
The $F_1$ improves from 0.313 for \textit{MiniInception} to 0.452 for \textit{MiniInception-Extended}, a relative improvement of 31\%.
Including the filtering step again provides a boost in performance leading an $F_1$ of 0.631.
This is a relative improvement of 10.8\% and demonstrates one of the key potential advantages of this deep learning approach that it benefits from increasing the training set size and annotating the training data is relatively easy as it required the labelling of a 2D image rather than a 3D point cloud (as is the case for the \textit{PFH-SVM} approach). Another approach would be to consider the use of synthetic data similar to~\cite{BARTH2017}.

\newcolumntype{s}{>{\centering\hsize=.5\hsize\arraybackslash}X}
\newcolumntype{Y}{>{\raggedright\hsize=.5\hsize\arraybackslash}X} 

\subsection{Experiment 3: Autonomous Harvesting} \label{subsec:experiment_3}

Results of the final harvesting experiment are presented in the following section. A video of the robotic harvester demonstrating the final experiment by performing autonomous harvesting of sweet peppers in a protected cropping environment is available at \url{http://bit.ly/experimental_results}.

\begin{figure}[htb]
	\centering
	\begin{tikzpicture}
	\begin{axis}[
	width=14cm,
	height=8cm,
	title={Harvesting Results},
	ybar,
	ymin=0,
	bar width=30,
	enlarge x limits=0.15,
		legend style={at={(0.98,0.95)},anchor=north east, legend columns=-1},	
	symbolic x coords={Pepper Detection,Peduncle Detection,Attachment Success,Harvest Success},
	xtick=data,
	nodes near coords={\pgfmathprintnumber\pgfplotspointmeta\%},
	xticklabel style={text width=3cm,align=center},
	nodes near coords align={vertical},
	]
	\addplot coordinates {(Pepper Detection,99) (Peduncle Detection,84) (Attachment Success,93) (Harvest Success, 76.5)};
	\addplot coordinates {(Pepper Detection,93) (Peduncle Detection,65) (Attachment Success,75) (Harvest Success, 47)};
	
	\legend{Modified,Unmodified}
	\end{axis}
	\end{tikzpicture}
	\caption{Harvesting rates for autonomous harvesting experiment within a protected cropping system. Sweet pepper detection success rates indicate whether a target sweet pepper was identified that fit within specified constraints (i.e enough points). Peduncle detection indicates the success of detecting a peduncle but doesn't reflect the accuracy which relates to the final harvesting success. Attachment success rates indicate whether the suction cup was placed onto the sweet pepper and created an effective grasp. Rates for ``Harvesting Success'' incorporate the performance from all previous stages including the detachment stage.} \label{fig:harvest_result_chart}
\end{figure}
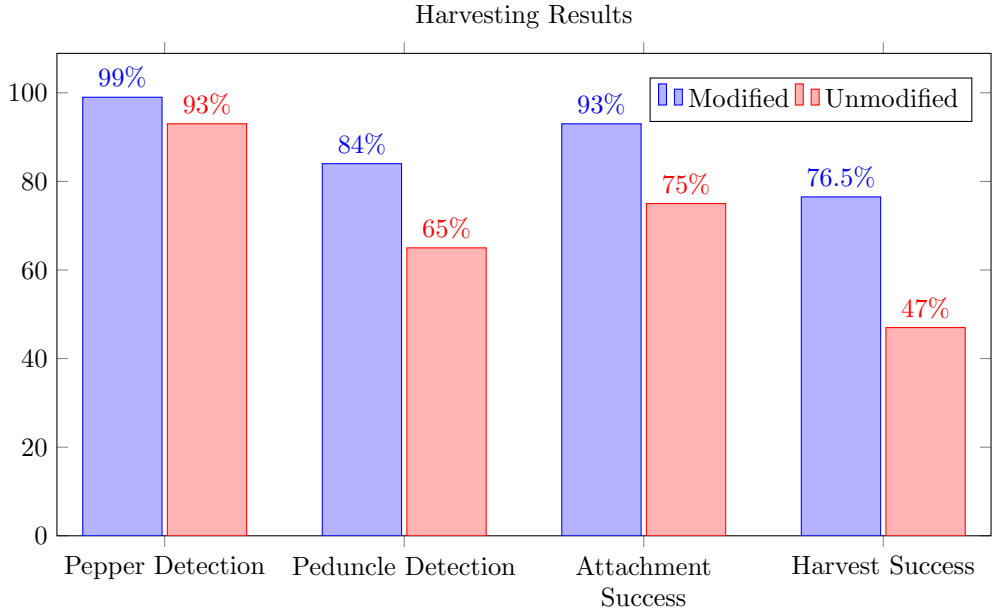

The success rates for the experiment are presented in Fig~\ref{fig:harvest_result_chart} for the unmodified and modified scenarios and are broken down into four different stages: sweet pepper segmentation, peduncle segmentation, attachment and overall harvest success rates. Out of the total sweet peppers, 76.5\% and 47\% were successfully harvested under the modified and unmodified scenarios respectively. Modifications were made, such as removing leaves or adjusting the sweet pepper pose. The overall successful harvesting rate reflects the performance of the detection, attachment and detachment stages. Table~\ref{tab:harvest_result} presents additional details for the harvesting experiment including the average number of attempts for each sweet pepper (1.9 for unmodified and 2.5 for modified) and the performance for the two different sweet pepper varieties. Arguably, one of the most challenging aspects of the harvesting process is the segmentation of peduncles---directly influencing whether a harvesting detachment can be executed. The results show that 84\% of peduncles were detected within the modified scenario and 65\% of the peduncles for the unmodified scenario. 

\begin{figure}[tbh]
	\centering
	\begin{subfigure}[b]{0.3\textwidth}\centering
		\includegraphics[width=\textwidth]{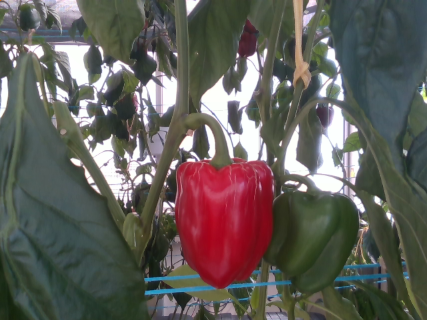}
	\end{subfigure}
	\begin{subfigure}[b]{0.3\textwidth}\centering
		\includegraphics[width=\textwidth]{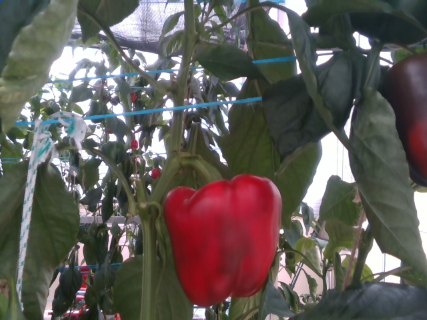}
	\end{subfigure}
	\begin{subfigure}[b]{0.3\textwidth}\centering
		\includegraphics[width=\textwidth]{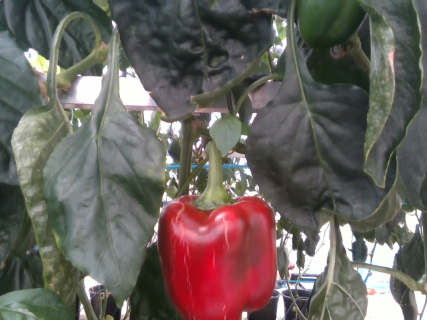}
	\end{subfigure}	
	\caption{Example images of sweet peppers taken during the final experiment from the robot's camera. A vertical offset is applied which shifts the sweet pepper down effectively centering the peduncle within the image aimed at improving the peduncle segmentation.}
	\label{fig:example_peppers}
\end{figure}

An example of images taken from the robot during the experiment are shown in Fig~\ref{fig:example_peppers}. These images show the perspective of the camera used to detect the sweet pepper and peduncle. It can be seen that the perspective is selected such that the peduncle is in the centre of the image maximising the number of peduncle pixels whilst still keeping the sweet pepper within the field of view. This camera perspective is determined using a previous estimate of the location of the target sweet pepper as described in Section~\ref{sec:detection}.

\begin{figure}[htb]
	\centering
\begin{subfigure}[b]{0.48\textwidth}\centering
	\begin{tikzpicture}
	\begin{axis}[
		width=\columnwidth,
	height=6cm,
		ybar,
	ymin=0,
	ymax=100,
	ylabel near ticks,
	ylabel={Damage Rate(\%)}
	bar width=20,
	enlarge x limits=0.15,
	legend style={at={(0.95,0.92)},anchor=north east, legend columns=-1},
	symbolic x coords={Minor Pepper Dmg,Minor Plant Dmg,Major Pepper Dmg,Major Plant Dmg},
	xtick=data,
	xticklabel style={font=\tiny, text width=1cm,align=center},
	nodes near coords={\pgfmathprintnumber\pgfplotspointmeta\%},
	nodes near coords align={vertical},
	]
	\addplot coordinates {(Minor Pepper Dmg,19) (Minor Plant Dmg, 18)(Major Pepper Dmg,7) (Major Plant Dmg,4)};
	\addplot coordinates {(Minor Pepper Dmg,12) (Minor Plant Dmg, 10) (Major Pepper Dmg,3) (Major Plant Dmg,3)};
	
	\legend{Modified,Unmodified}
	\end{axis}
	\end{tikzpicture}
	\caption{} 		
	\label{fig:harvest_dmg_result_chart}
\end{subfigure}
\begin{subfigure}[b]{0.48\textwidth}\centering
	\raisebox{0mm}{\begin{tikzpicture}
	\begin{axis}[
		width=\columnwidth,
	height=6cm,
		ybar,
	ymin=0,
	ymax=20,
	bar width=25,
	ylabel near ticks,
	ylabel={Time($s$)},
	enlarge x limits=0.1,
		symbolic x coords={Pepper Detection, Grasp Selection, Peduncle Detection, Attach, Detach, Place},
	xtick=data,
	xticklabel style={font=\tiny,text width=1cm,align=center},
	nodes near coords align={vertical},
	]
	
	\addplot+[error bars/.cd,
	y dir=both,y explicit] coordinates {
	 (Pepper Detection, 4.3) +-(1.2,1.2)
	 (Grasp Selection, 0.9)+-(0.5,0.5)
	 (Peduncle Detection, 1.4) +-(2.2,2.2)
	 (Attach, 6.7) +-(4.7,4.7)
	 (Detach, 14.5) +-(2.9,2.9)
	 (Place, 9.2) +-(2.4,2.4)
	};
		
	\end{axis}
	\end{tikzpicture}}
	\caption{} 
	\label{fig:harvest_timing_result}
\end{subfigure}
\caption{(a) Sweet pepper and plant damage rates for autonomous harvesting. (b) Average time for each stage of the harvesting process ($N=68$). Error bars indicate one standard deviation from the mean.}
\end{figure}
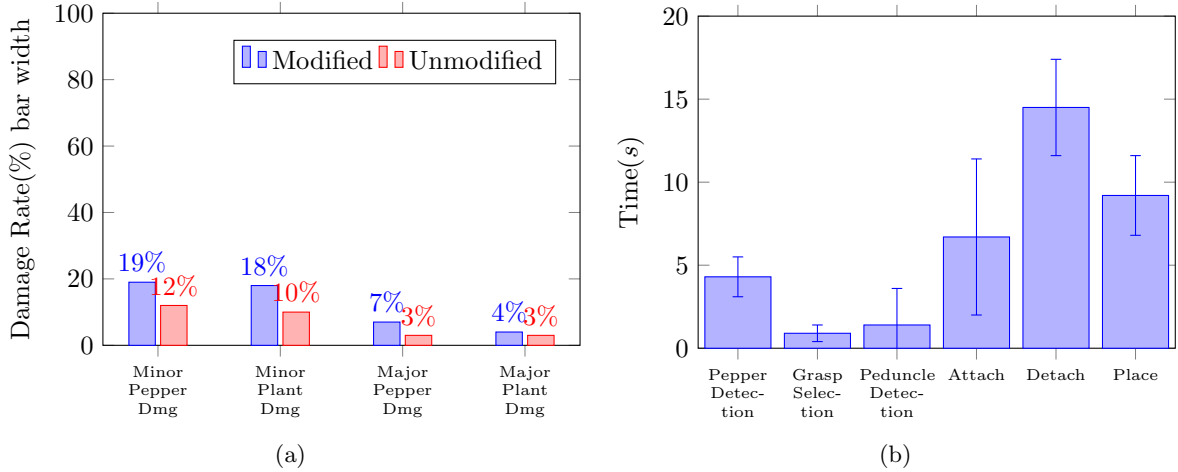

Within a protected cropping system, damage to the plant can lead to a loss in yield of that plant due to disease, reduced growth or worse case, death from the damage. It is therefore critical that the harvesting method achieves as minimal damage to the plant as possible. During the final experiment, any damage to sweet peppers or the plant was recorded. Fig~\ref{fig:harvest_dmg_result_chart} shows the damage rates for the modified and unmodified experiments, separated into major and minor damage to either the sweet pepper or plant. Results show that 5 (7\%) sweet peppers suffered major damage (see Fig~\ref{fig:example_sweet_pepper_dmg} for an example) and 3 (4\%) plant/stems had major damage for the modified scenario. Alternative designs for the cutting tool, such as the addition of a guard, could reduce damage to both the plant and fruit.

An analysis of the timing for each stage of the harvesting process is illustrated in Fig~\ref{fig:harvest_timing_result} and presented in Table \ref{tab:timing_result}. The average time to detach a sweet pepper from the plant was 36.9 ($\pm 6.4$) seconds where the most time was spent on the detachment stage (14.5 ($\pm 2.9$) seconds). The detachment stage was the most time consuming as it involved the cutting motion which was executed with a slow end effector velocity to ensure the oscillating cutting blade successfully severed the peduncle. This detachment time could be reduced if a more powerful oscillating tool was used. The other major time consuming stages are the attachment and placement stages as these also require motion of the robot arm. 

Unfortunately the placement stage was very inefficient in time (9.2 ($\pm 2.4$) seconds) as it was discovered after the experiment that the path planning algorithm was taking a large amount of time (including multiple planning attempts) to find a path to the packing crate due to a poor choice of predefined waypoints. This problem could easily be resolved in future work by choosing better waypoints for the packing crate ensuring the path planning algorithm doesn't waste a significant amount of time re-planning.

\begin{table}[htb]
	\centering
	\caption{Success rates for autonomous harvesting experiment}
	\label{tab:harvest_result}
	\begin{tabularx}{\columnwidth}{ls|s|s|s}
		\toprule
		                         & Unmodified     & Modified       & Mercuno Variety (Modified) & Ducati Variety (Modified)   \\ \midrule
		Avg Attempts (total)     & 1.9    & 2.5     & 2.9     &  2.3  \\[5pt]
		 Pepper Detection      & 93\%   &  99\%   &  95\%   & 96\% \\[5pt]
		Peduncle Detection       &  65\%  &  84\%   &  86\%   &  81\% \\[5pt]
		Attachment Success       & 75\%   &  93\%   &  95\%   &  90\% \\[5pt]
		Overall Harvest Success  &  47\%  &  76.5\% &  81\%   &  71\% \\[5pt]
		\bottomrule
	\end{tabularx}
\end{table}

\begin{table}[htb]
	\centering
	\caption{Execution Times for Harvesting Stages}
	\label{tab:timing_result}
	\begin{tabularx}{0.5\columnwidth}{l|s}
		\toprule
		\textbf{Harvesting stage}   & \textbf{Average time in seconds (std dev)}  \\\midrule
		Sweet Pepper Detection   & 4.3  ($\pm 1.2$)     \\[5pt]
		Grasp Selection       & 0.9  ($\pm 0.5$)  \\[5pt]
		Peduncle Detection    & 1.4  ($\pm 2.2$)  \\[5pt]
		Attachment            & 6.7  ($\pm 4.7$)  \\[5pt]
		Detachment            & 14.5 ($\pm 2.9$)  \\[5pt]
		Placement             & 9.2  ($\pm 2.4$)  \\[5pt]\midrule
		\textbf{Total}        & \textbf{36.9 ($\pm 6.4$)}  \\[5pt]
		\bottomrule
	\end{tabularx}
\end{table}

\subsubsection{Failure Analysis}

\begin{figure}[htb]
	\centering
	\begin{tikzpicture}
	\begin{axis}[
		width=\columnwidth,
	height=6cm,
	title={Harvesting Failure Conditions },
	ybar,
	ymin=0,
	ymax=100,
	bar width=15,
	enlarge x limits=0.1,
	legend style={at={(0.95,0.92)},anchor=north east, legend columns=-1},
	symbolic x coords={(a), (b), (c), (d), (e), (f), (g), (h), (i)},
	xtick=data,
	xticklabel style={text width=1cm,align=center},
	nodes near coords={\pgfmathprintnumber\pgfplotspointmeta\%},
	nodes near coords align={vertical},
	]
	
	\addplot coordinates {
	 ((a), 28)((b), 23) ((c), 22) ((d), 11)((e), 6) ((f), 7) ((g), 5) ((h), 5) ((i), 2)
	};
	
		\addplot coordinates {
	 ((a), 30)((b), 20) ((c), 24)((d), 10)((e), 9) ((f), 7)((g), 5) ((h), 4) ((i), 2)
	};
		
	\legend{Modified,Unmodified}
	\end{axis}
	\end{tikzpicture}
	
	\caption{Harvesting Failure conditions (N=127 for modified, N=99 for unmodified). The different failure cases include:
		a) Peduncle Not detected 
		b) Peduncle partially cut 
		c) Peduncle moved     		
		d) Difficult peduncle
		e) Path planning failure          
		f) Obstruction of peduncle        
		g) Difficult sweet pepper         
		h) Obstruction of sweet pepper    
		i) Attachment failure.}
	\label{fig:harvest_failures}
\end{figure}
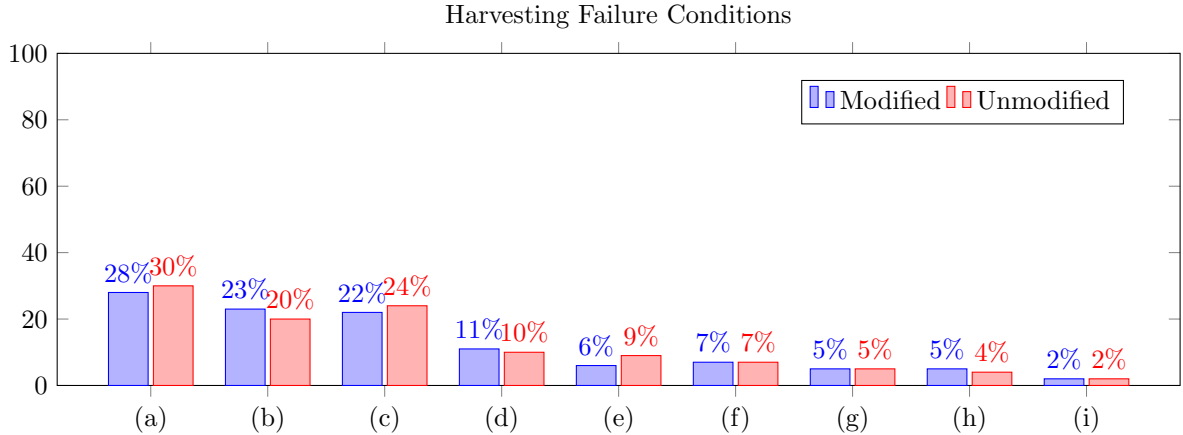

It is important for future work to understand what are the major causes of harvesting failures within the system proposed. Therefore, a failure analysis was conducted to highlight what are some of the reoccurring modes of failure. Each attempt that was performed during the experiment that did not lead to a successful harvest was used within the failure analysis. Using notes from the field trial, the modes of failure were broken down into the following categories:
\begin{enumerate}[parsep=0pt,label=\alph*)]
	\item Peduncle not detected - the peduncle was challenging to detect
	\item Peduncle partially cut - the cutting tool partially severed the peduncle
	\item Peduncle moved - the peduncle shifted out of the way during the cutting action      		
	\item Difficult peduncle - the peduncle was abnormal in size or shape making it difficult for cutting
	\item Path planning failure - the manipulator was unable to plan the grasping or cutting action
	\item Obstruction of peduncle - the peduncle was fully occluded         
	\item Difficult sweet pepper - the sweet pepper was abnormal in size or shape making it difficult for grasping          
	\item Obstruction of sweet pepper - the sweet pepper was fully occluded 
	\item Attachment failure - the gripper was unable to grasp the sweet pepper
\end{enumerate}
Each failure was recorded and the rates at which they occurred are presented in Fig~\ref{fig:harvest_failures}. The results show that the most frequent failure mode was (a) no peduncle detected, occurring 28\% (30\% for unmodified scenario) of the time. The failure modes (b) and (c) also represented a significant portion of the failure cases and show that detachment stage is challenging due a combination of inaccuracies of the peduncle detection and dynamic movement of peduncles during cutting.

\section{Discussion \& Conclusion} \label{sec:discussion_and_conclusion}

This paper describes an autonomous crop harvesting system that achieves impressive harvesting results in a real protected cropping environment. Automating the currently manual horticultural harvesting task has been a long sought after goal with relatively slow development over the past few decades \citep{Bac2014} and with little commercial impact \citep{Shamshiri2018}. However, in this paper we have demonstrated a robotic system that approaches commercial viability. We highlight two key challenges for successful autonomous harvesting: 1) perception of the crop and environment; and 2) manipulation of the crop. This paper makes three key contributions: 

\begin{itemize}
	\item A proven in-field robotic harvesting system that achieves a harvesting success rate of 76.5\% in a modified scenario,
	\item a in-depth analysis of the perception and harvesting field trials of the robotic harvester,
	\item and a novel method for peduncle segmentation using an efficient deep convolutional neural network in conjunction with 3D post-filtering.
\end{itemize}

The results demonstrate that visual detection of the crop and peduncle can be achieved in very difficult situations including challenging lighting conditions and with highly similar visual appearance to the background. We demonstrate that the combination of efficient deep learning models and 3D filtering reduce problems with estimating the critical cutting location within the harvesting process. Finally, the system is demonstrated using a custom end-effector that uses both a suction gripper and oscillating blade to successfully remove sweet peppers in a protected cropping environment. The presented harvesting system achieved a 76.5\% success rate (within a modified scenario) which improves upon our prior work which achieved 58\% and related sweet pepper harvesting work which achieved 33\%~\citep{Bac2017} (which has some differences in the cropping system). Despite these improvements over the state of the art, a number of issues were made apparent throughout our experiments. These are broken down into the perception, harvesting and hardware design issues discussed below.

The perception and planning systems have been shown to perform well for our specific problem. However, further advancements are necessary for a general system. To achieve a generic visual crop detection system suitable for a range of crops, we believe a fast learning system capable of high detection accuracy that can be trained from a small number of training images is required. This is a challenging task which remains unsolved.

The harvesting cycle time is currently slow (approximately 37 seconds), and reducing this cycle time would be of high priority for future work. One of the most time consuming processes is the detachment step, as the end-effector is moved at a slow velocity to ensure the peduncle is fully severed. This can be addressed by improving the cutting rate of the oscillating tool such as: increasing its power; or investigating different blade shapes (which could also improve the robustness of the cutting action under position uncertainty). Improvements to the cutting rate enable the end-effector velocity to be increased during the cutting action.

A common detachment failure case was either the blade partially cutting the peduncle or the peduncle moved out of the way during the cutting action. Improving the deep learning system by increasing the training data is one possible method to improve detachment reliability. Adding visual servoing methods could also increase the robustness to changes in the environment such as when the peduncle shifts during the cutting action. 
Another challenge remaining with the methods presented in this paper are in improving reliable attachment and detachment for the unmodified scenarios which include challenging visual and physical obstructions from leaves and the surrounding plant. One potential way to improve the system would be to use active perception (moving to see) methods which decide on how to move the camera to improve the view of the crop that may be highly occluded from leaves.

Protected cropping systems in tropical climates such as Northern Australia can differ to other international greenhouse systems with respect to trellising and potting methods. However, the underlying plant structure such as leaves, stems and sweet peppers are very similar including their physical and visual appearance. Under these assumptions we believe the work demonstrated could easily be applied to protected cropping systems in sub-tropical climates which feature vertical trellising and pipe rail heating systems.

The novel contributions of this work have resulted in considerable and extremely encouraging improvements in sweet pepper picking success rates compared with the state-of-the-art. We believe that continuing to build on the system presented in this paper will result in further meaningful progress towards making an impact in the horticulture industry through a commercially viable system. The methods presented in this paper may also be applied to a range of other high-value horticultural crops, and provides steps towards the ultimate goal of fully autonomous and reliable crop management systems to reduce labour costs, maximise the quality of produce, and ultimately improve the sustainability of farming enterprises.
\subsubsection*{Acknowledgments}
This work was supported by the Strategic Investment in Farm Robotics (SIFR) program---a initiative co-funded by QUT and the Department of Agriculture and Fisheries of the Queensland Government,  Queensland, Australia.
Contributions to the work have also been supported by QUT's Institute for Future Environments. 

We would like to thank Dr Elio Jovicich and Heidi Wiggenhauser of Queensland DAF for their help, support, and feedback towards organising and conducting the field trial of the robotic harvesting platform.

\bibliographystyle{apalike}
\bibliography{./bibs/mccool_bib,./bibs/jfrRefs}

\begin{thebibliography}{}

\bibitem[ABARE, 2014]{ABARE2014}
ABARE (2014).
\newblock Australian vegetable growing farms: An economic survey, 2012-13 and
  2013-14.
\newblock Research report, Australian Bureau of Agricultural and Resource
  Economics (ABARE).

\bibitem[Bac et~al., 2017]{Bac2017}
Bac, C.~W., Hemming, J., van Tuijl, B.~A., Barth, R., Wais, E., and van Henten,
  E.~J. (2017).
\newblock {Performance Evaluation of a Harvesting Robot for Sweet Pepper}.
\newblock {\em Journal of Field Robotics}, 34(6):1123--1139.

\bibitem[Bac et~al., 2014]{Bac2014}
Bac, C.~W., van Henten, E.~J., Hemming, J., and Edan, Y. (2014).
\newblock {Harvesting Robots for High-value Crops: State-of-the-art Review and
  Challenges Ahead}.
\newblock {\em Journal of Field Robotics}, pages n/a--n/a.

\bibitem[Bachche and Oka, 2013]{Bachche2013}
Bachche, S. and Oka, K. (2013).
\newblock {Performance Testing of Thermal Cutting Systems for Sweet Pepper
  Harvesting Robot in Greenhouse Horticulture}.
\newblock {\em Journal of System Design and Dynamics}, 7(1):36--51.

\bibitem[Baeten et~al., 2008]{Baeten2008}
Baeten, J., Donn{\'e}, K., Boedrij, S., Beckers, W., and Claesen, E. (2008).
\newblock Autonomous fruit picking machine: A robotic apple harvester.
\newblock In {\em Field and Service Robotics}, pages 531--539. Springer.

\bibitem[Barth et~al., 2016]{Barth2016}
Barth, R., Hemming, J., and van Henten, E.~J. (2016).
\newblock {Design of an eye-in-hand sensing and servo control framework for
  harvesting robotics in dense vegetation}.
\newblock {\em Biosystems Engineering}.

\bibitem[Barth et~al., 2017]{BARTH2017}
Barth, R., IJsselmuiden, J., Hemming, J., and Henten, E.~V. (2017).
\newblock Synthetic bootstrapping of convolutional neural networks for semantic
  plant part segmentation.
\newblock {\em Computers and Electronics in Agriculture}.

\bibitem[Baur et~al., 2014]{Baur2014}
Baur, J., Sch{\"{u}}tz, C., and Pfaff, J. (2014).
\newblock {Path Planning for a Fruit Picking Manipulator}.
\newblock In {\em International Conference of Agricultural Engineering}.

\bibitem[Beeson and Ames, 2015]{Beeson2015}
Beeson, P. and Ames, B. (2015).
\newblock Trac-ik: An open-source library for improved solving of generic
  inverse kinematics.
\newblock In {\em Humanoid Robots (Humanoids), 2015 IEEE-RAS 15th International
  Conference on}, pages 928--935. IEEE.

\bibitem[Bhattacharjee et~al., 2014]{Bhattacharjee2014}
Bhattacharjee, T., Grice, P.~M., Kapusta, A., Killpack, M.~D., Park, D., and
  Kemp, C.~C. (2014).
\newblock A robotic system for reaching in dense clutter that integrates model
  predictive control, learning, haptic mapping, and planning.
\newblock In {\em IEEE/RSJ International Conference on Intelligent Robots and
  Systems (IROS 2014)-3rd Workshop on Robots in Clutter: Perception and
  Interaction in Clutter}, pages 14--18.

\bibitem[Blanes et~al., 2011]{Blanes2011}
Blanes, C., Mellado, M., Ortiz, C., and Valera, A. (2011).
\newblock Review. technologies for robot grippers in pick and place operations
  for fresh fruits and vegetables.
\newblock {\em Spanish Journal of Agricultural Research}, 9(4):1130--1141.

\bibitem[Blasco et~al., 2003]{blasco2003machine}
Blasco, J., Aleixos, N., and Molt{\'o}, E. (2003).
\newblock Machine vision system for automatic quality grading of fruit.
\newblock {\em Biosystems Engineering}, 85(4):415--423.

\bibitem[Bohren and Cousins, 2010]{smach}
Bohren, J. and Cousins, S. (2010).
\newblock The smach high-level executive [ros news].
\newblock {\em IEEE Robotics \& Automation Magazine}, 17(4):18--20.

\bibitem[Bontsema et~al., 2014]{Bontsema2014}
Bontsema, J., Hemming, J., Saeys, E. P.~W., Edan, Y., Shapiro, A., Ho\v{c}evar,
  M., Hellstr\"{o}m, T., Oberti, R., Armada, M., Ulbrich, H., Baur, J.,
  Debilde, B., Best, S., Evain, S., M\"{u}nzenmaier, A., and Ringdahl, O.
  (2014).
\newblock {CROPS: high tech agricultural robots}.
\newblock In {\em International Conference of Agricultural Engineering}, pages
  6--10.

\bibitem[Bulanon and Kataoka, 2010a]{Bulanon2010}
Bulanon, D. and Kataoka, T. (2010a).
\newblock {Fruit detection system and an end effector for robotic harvesting of
  Fuji apples}.
\newblock {\em Agricultural Engineering International: CIGR}, 12(1):203--210.

\bibitem[Bulanon and Kataoka, 2010b]{Bulanon2010a}
Bulanon, D.~M. and Kataoka, T. (2010b).
\newblock {A Fruit Detection System and an End Effector for Robotic Harvesting
  of Fuji Apples}.
\newblock XII:1--14.

\bibitem[Cubero et~al., 2014]{cubero2014}
Cubero, S., Diago, M.~P., Blasco, J., Tard{\'a}guila, J., Mill{\'a}n, B., and
  Aleixos, N. (2014).
\newblock A new method for pedicel/peduncle detection and size assessment of
  grapevine berries and other fruits by image analysis.
\newblock {\em Biosystems engineering}, 117:62--72.

\bibitem[De-An et~al., 2011]{De-An2011}
De-An, Z., Jidong, L., Wei, J., Ying, Z., and Yu, C. (2011).
\newblock {Design and control of an apple harvesting robot}.
\newblock {\em Biosystems engineering}.

\bibitem[Gongal et~al., 2015]{Gongal2015}
Gongal, A., Amatya, S., Karkee, M., Zhang, Q., and Lewis, K. (2015).
\newblock {Sensors and systems for fruit detection and localization: A review}.
\newblock {\em Computers and Electronics in Agriculture}, 116:8--19.

\bibitem[Gongal et~al., 2016]{gongal2016apple}
Gongal, A., Silwal, A., Amatya, S., Karkee, M., Zhang, Q., and Lewis, K.
  (2016).
\newblock Apple crop-load estimation with over-the-row machine vision system.
\newblock {\em Computers and Electronics in Agriculture}, 120:26--35.

\bibitem[Han et~al., 2012]{Han2012}
Han, K.-S., Kim, S.-C., Lee, Y.-B., Kim, S.-C., Im, D.-H., Choi, H.-K., and
  Hwang, H. (2012).
\newblock Strawberry harvesting robot for bench-type cultivation.
\newblock {\em Journal of Biosystems Engineering}, 37(1):65--74.

\bibitem[Hannan and Burks, 2004]{Hannan2004}
Hannan, M. and Burks, T. (2004).
\newblock {Current developments in automated citrus harvesting}.
\newblock {\em 2004 ASAE Annual Meeting}.

\bibitem[Hayashi et~al., 2010]{Hayashi2010}
Hayashi, S., Shigematsu, K., Yamamoto, S., Kobayashi, K., Kohno, Y., Kamata,
  J., and Kurita, M. (2010).
\newblock {Evaluation of a strawberry-harvesting robot in a field test}.
\newblock {\em Biosystems Engineering}, 105(2):160--171.

\bibitem[Hemming et~al., 2014]{Hemming2014a}
Hemming, J., Bac, C., van Tuijl, B., Barth, R., Bontsema, J., Pekkeriet, E.,
  and van Henten, E. (2014).
\newblock {A robot for harvesting sweet-pepper in greenhouses}.
\newblock In {\em Proceedings of the International Conference of Agricultural
  Engineering}.

\bibitem[Hemming et~al., 2013]{Hemming2013b}
Hemming, J., van Tuijl, B., Bac, W., and Barth, R. (2013).
\newblock {Test report of modules of harvester, including suggestions for
  revision and improvement}.
\newblock 33.

\bibitem[Hung et~al., 2013]{Hung:2013aa}
Hung, C., Nieto, J., Taylor, Z., Underwood, J., and Sukkarieh, S. (2013).
\newblock Orchard fruit segmentation using multi-spectral feature learning.
\newblock In {\em IEEE/RSJ International Conference on Intelligent Robots and
  Systems}, pages 5314--5320.

\bibitem[Ilievski et~al., 2011]{Ilievski2011}
Ilievski, F., Mazzeo, A.~D., Shepherd, R.~F., Chen, X., and Whitesides, G.~M.
  (2011).
\newblock Soft robotics for chemists.
\newblock {\em Angewandte Chemie International Edition}, 50(8):1890--1895.

\bibitem[Inc., 2016]{SoftRobotics2016}
Inc., S.~R. (2016).
\newblock Soft robotics enabling new markets in automation.

\bibitem[Jain et~al., 2013]{Jain2013}
Jain, A., Killpack, M.~D., Edsinger, A., and Kemp, C.~C. (2013).
\newblock Reaching in clutter with whole-arm tactile sensing.
\newblock {\em The International Journal of Robotics Research}, page
  0278364912471865.

\bibitem[Karaman and Frazzoli, 2011]{rrtstar}
Karaman, S. and Frazzoli, E. (2011).
\newblock Sampling-based algorithms for optimal motion planning.
\newblock {\em The International Journal of Robotics Research}, 30(7):846--894.

\bibitem[Killpack et~al., 2015]{Killpack2015}
Killpack, M., Kapusta, A., and Kemp, C. (2015).
\newblock {Model predictive control for fast reaching in clutter}.
\newblock {\em Autonomous Robots}.

\bibitem[Kitamura and Oka, 2005]{Kitamura2005}
Kitamura, S. and Oka, K. (2005).
\newblock Recognition and cutting system of sweet pepper for picking robot in
  greenhouse horticulture.
\newblock In {\em Mechatronics and Automation, 2005 IEEE International
  Conference}, volume~4, pages 1807--1812. IEEE.

\bibitem[Kitamura et~al., 2008]{Kitamura2008}
Kitamura, S., Oka, K., Ikutomo, K., Kimura, Y., and Taniguchi, Y. (2008).
\newblock {A distinction method for fruit of sweet pepper using reflection of
  LED light}.
\newblock {\em 2008 SICE Annual Conference}, 1:491--494.

\bibitem[Kondo et~al., 2011]{Kondoetall2011}
Kondo, N., Monta, M., and Noguchi, N., editors (2011).
\newblock {\em Agricultural Robots: Mechanisms and Practice}.
\newblock Trans Pacific Press.

\bibitem[Kondo et~al., 1996]{Kondo1996}
Kondo, N., Nishitsuji, Y., Ling, P.~P., and Ting, K.~C. (1996).
\newblock Visual feedback guided robotic cherry tomato harvesting.
\newblock {\em Transactions of the ASAE}, 39(6):2331--2338.

\bibitem[Kondo et~al., 2010]{Kondo2010}
Kondo, N., Yata, K., Iida, M., Shiigi, T., Monta, M., Kurita, M., and Omori, H.
  (2010).
\newblock Development of an end-effector for a tomato cluster harvesting robot.
\newblock {\em Engineering in Agriculture, Environment and Food}, 3(1):20--24.

\bibitem[Lehnert et~al., 2015]{Lehnert2015}
Lehnert, C., Perez, T., and McCool, C. (2015).
\newblock {Optimisation-based Design of a Manipulator for Harvesting Capsicum}.

\bibitem[Lehnert et~al., 2016]{Lehnert2016}
Lehnert, C., Sa, I., Mccool, C., Upcroft, B., and Perez, T. (2016).
\newblock {Sweet Pepper Pose Detection and Grasping for Automated Crop
  Harvesting}.
\newblock In {\em International Conference on Robotics and Automation}.

\bibitem[Lehnert et~al., 2017]{Lehnert2017}
Lehnert, C.~F., English, A., McCool, C., Tow, A.~W., and Perez, T. (2017).
\newblock Autonomous sweet pepper harvesting for protected cropping systems.
\newblock {\em IEEE Robotics and Automation Letters}, 2(2):872--879.

\bibitem[Ling et~al., 2004]{Ling2005}
Ling, P.~P., Ehsani, R., Ting, K., Chi, Y.-T., Ramalingam, N., Klingman, M.~H.,
  and Draper, C. (2004).
\newblock Sensing and end-effector for a robotic tomato harvester.
\newblock In {\em 2004 ASAE Annual Meeting}, page~1. American Society of
  Agricultural and Biological Engineers.

\bibitem[McCool et~al., 2017]{McCool17_1}
McCool, C., Perez, T., and Upcroft, B. (2017).
\newblock Mixture of lightweight deep convolutional neural networks: applied to
  agricultural robotics.
\newblock {\em IEEE Robotics and Automation Letters}.

\bibitem[McCool et~al., 2016]{McCool:2016aa}
McCool, C., Sa, I., Dayoub, F., Lehnert, C., Perez, T., and Upcroft, B. (2016).
\newblock {Visual Detection of Occluded Crop: for automated harvesting}.
\newblock In {\em The International Conference on Robotics and Automation}.

\bibitem[Mehta and Burks, 2014]{Mehta2014}
Mehta, S. and Burks, T. (2014).
\newblock {Vision-based control of robotic manipulator for citrus harvesting}.
\newblock {\em Computers and Electronics in Agriculture}, 102:146--158.

\bibitem[Milioto and Stachniss, 2018]{milioto2018arxiv}
Milioto, A. and Stachniss, C. (2018).
\newblock {Bonnet: An Open-Source Training and Deployment Framework for
  Semantic Segmentation in Robotics using CNNs}.
\newblock {\em ArXiv e-prints}.

\bibitem[Monkman et~al., 2007]{Monkman2007}
Monkman, G., Hesse, S., Steinmann, R., and Schunk, H. (2007).
\newblock {\em {Robot grippers}}.

\bibitem[Nguyen et~al., 2014]{Nguyen2014}
Nguyen, T.~T., Vandevoorde, K., Kayacan, E., De~Baerdemaeker, J., and Saeys, W.
  (2014).
\newblock Apple detection algorithm for robotic harvesting using a rgb-d
  camera.
\newblock In {\em International Conference of Agricultural Engineering, Zurich,
  Switzerland}.

\bibitem[Nuske et~al., 2014]{Nuske:2014aa}
Nuske, S., Wilshusen, K., Achar, S., Yoder, L., Narasimhan, S., and Singh, S.
  (2014).
\newblock Automated visual yield estimation in vineyards.
\newblock {\em Journal of Field Robotics}, 31(5):837--860.

\bibitem[Nuske et~al., 2011]{Nuske_2011_6891}
Nuske, S.~T., Achar, S., Bates, T., Narasimhan, S.~G., and Singh, S. (2011).
\newblock Yield estimation in vineyards by visual grape detection.
\newblock In {\em Proceedings of the 2011 IEEE/RSJ International Conference on
  Intelligent Robots and Systems (IROS '11)}.

\bibitem[Rahnemoonfar and Sheppard, 2016]{Rahnemoonfar16_1}
Rahnemoonfar, M. and Sheppard, C. (2016).
\newblock Deep count: Fruit counting based on deep simulated learning.
\newblock {\em Sensors}.

\bibitem[Ruiz et~al., 1996]{Ruiz:1996aa}
Ruiz, L.~A., Molt{\'o}, E., Juste, F., Pl{\'a}, F., and Valiente, R. (1996).
\newblock Location and characterization of the stem--calyx area on oranges by
  computer vision.
\newblock {\em Journal of Agricultural Engineering Research}, 64(3):165--172.

\bibitem[Rusu et~al., 2009]{Rusu2009-kl}
Rusu, R.~B., Blodow, N., and Beetz, M. (2009).
\newblock {Fast Point Feature Histograms ({FPFH}) for {3D} registration}.
\newblock In {\em 2009 {IEEE} International Conference on Robotics and
  Automation}, pages 3212--3217.

\bibitem[Sa et~al., 2016]{Sa16_1}
Sa, I., Ge, Z., Dayoub, F., Upcroft, B., Perez, T., and McCool, C. (2016).
\newblock Deepfruits: A fruit detection system using deep neural networks.
\newblock {\em Sensors}.

\bibitem[Sa et~al., 2017]{Sa17_1}
Sa, I., Lehnert, C., English, A., McCool, C., Dayoub, F., Upcroft, B., and
  Perez, T. (2017).
\newblock {Peduncle detection of sweet pepper for autonomous crop
  harvesting---Combined Color and 3-D Information}.
\newblock {\em IEEE Robotics and Automation Letters}, 2(2):765--772.

\bibitem[Scarfe et~al., 2009]{Scarfe2000}
Scarfe, A.~J., Flemmer, R.~C., Bakker, H., and Flemmer, C.~L. (2009).
\newblock Development of an autonomous kiwifruit picking robot.
\newblock In {\em Autonomous Robots and Agents, 2009. ICARA 2009. 4th
  International Conference on}, pages 380--384. IEEE.

\bibitem[Schuetz et~al., 2015]{Schuetz2015}
Schuetz, C., Baur, J., Pfaff, J., Buschmann, T., and Ulbrich, H. (2015).
\newblock Evaluation of a direct optimization method for trajectory planning of
  a 9-dof redundant fruit-picking manipulator.
\newblock In {\em Robotics and Automation (ICRA), 2015 IEEE International
  Conference on}, pages 2660--2666. IEEE.

\bibitem[Shamshiri et~al., 2018]{Shamshiri2018}
Shamshiri, R.~R., Weltzien, C., Hameed, I.~A., Yule, I.~J., Grift, T.~E.,
  Balasundram, S.~K., Pitonakova, L., Ahmad, D., and Chowdhary, G. (2018).
\newblock {Research and development in agricultural robotics: A perspective of
  digital farming}.
\newblock {\em International Journal of Agricultural and Biological
  Engineering}, 11(4):1--14.

\bibitem[Sucan and Chitta, 2016]{moveit!}
Sucan, I.~A. and Chitta, S. (2016).
\newblock Moveit!

\bibitem[{\c{S}}ucan et~al., 2012]{ompl}
{\c{S}}ucan, I.~A., Moll, M., and Kavraki, L.~E. (2012).
\newblock The {O}pen {M}otion {P}lanning {L}ibrary.
\newblock {\em {IEEE} Robotics \& Automation Magazine}, 19(4):72--82.
\newblock \url{http://ompl.kavrakilab.org}.

\bibitem[van Henten and Hemming, 2002]{Henten2002}
van Henten, E. and Hemming, J. (2002).
\newblock {An autonomous robot for harvesting cucumbers in greenhouses}.
\newblock {\em Autonomous Robots}, pages 241--258.

\bibitem[{Van Henten} et~al., 2003]{VanHenten2003}
{Van Henten}, E., {Van Tuijl}, B., Hemming, J., Kornet, J., Bontsema, J., and
  {Van Os}, E. (2003).
\newblock {Field Test of an Autonomous Cucumber Picking Robot}.
\newblock {\em Biosystems Engineering}, 86(3):305--313.

\bibitem[{Van Henten} et~al., 2009]{VanHenten2009}
{Van Henten}, E.~J., {Van't Slot}, D.~a., Hol, C. W.~J., and {Van
  Willigenburg}, L.~G. (2009).
\newblock {Optimal manipulator design for a cucumber harvesting robot}.
\newblock {\em Computers and Electronics in Agriculture}, 65(2):247--257.

\bibitem[Wang et~al., 2012]{Wang_2012_7240}
Wang, Q., Nuske, S.~T., Bergerman, M., and Singh, S. (2012).
\newblock Automated crop yield estimation for apple orchards.
\newblock In {\em 13th Internation Symposium on Experimental Robotics (ISER
  2012)}, number CMU-RI-TR-.

\bibitem[Yamamoto et~al., 2014]{Yamamoto:2014aa}
Yamamoto, K., Guo, W., Yoshioka, Y., and Ninomiya, S. (2014).
\newblock {On Plant Detection of Intact Tomato Fruits Using Image Analysis and
  Machine Learning Methods}.
\newblock {\em Sensors}.

\end{thebibliography}

\end{document}